\documentclass{article}

\PassOptionsToPackage{numbers, compress}{natbib}
\usepackage[sglblindworkshop, final]{_neurips_2025}
\workshoptitle{MLxOR: Mathematical Foundations and Operational Integration of Machine Learning for Uncertainty-Aware Decision-Making}

\usepackage[utf8]{inputenc} 
\usepackage[T1]{fontenc}    
\usepackage{hyperref}       
\usepackage{url}            
\usepackage{booktabs}       
\usepackage{nicefrac}       
\usepackage{microtype}      
\usepackage{xcolor}         

\usepackage{amsmath,amsfonts,amssymb,amsthm}
\usepackage[toc]{appendix}
\usepackage{titletoc}
\usepackage{tocloft}

\usepackage{placeins}  
\usepackage{graphicx}
\usepackage{subcaption}
\usepackage{wrapfig}
\usepackage{ragged2e}
\usepackage{algorithm}
\usepackage{algorithmic}
\usepackage{threeparttable}
\usepackage{multicol,multirow}

\hypersetup{colorlinks=true,allcolors=blue}
\title{Uncertainty Estimation using Variance-Gated Distributions}
\workshoptitle{ML$\times$OR Workshop: Mathematical Foundations and Operational Integration of Machine Learning for Uncertainty-Aware Decision-Making} 
\author{%
  H.~Martin Gillis\thanks{Equal contributions.} \\
  Faculty of Computer Science \\
  Dalhousie University \\
  Halifax, NS Canada \\
  \texttt{martin.gillis@dal.ca} \\
  \And
  Isaac Xu\footnotemark[1] \\
  Faculty of Computer Science \\
  Dalhousie University \\
  Halifax, NS Canada \\
  \texttt{isaac.xu@dal.ca} \\
  \And
  Thomas Trappenberg\thanks{Corresponding author.} \\
  Faculty of Computer Science \\
  Dalhousie University \\
  Halifax, NS Canada \\
  \texttt{tt@cs.dal.ca}
}
\begin{document}
\maketitle

\begin{abstract}
Evaluation of per-sample uncertainty quantification from neural networks is essential for decision-making involving high-risk applications.
A common approach is to use the predictive distribution from Bayesian or approximation models and decompose the corresponding predictive uncertainty into epistemic (model-related) and aleatoric (data-related) components.
However, additive decomposition has recently been questioned.
In this work, we propose an intuitive framework for uncertainty estimation and decomposition based on the signal-to-noise ratio of class probability distributions across different model predictions.
We introduce a variance-gated measure that scales predictions by a confidence factor derived from ensembles. We use this measure to discuss the existence of a collapse in the diversity of committee machines. 
\end{abstract}

\section{Introduction} \label{sec:introduction}
High-risk decision-making using machine learning systems for applications such as healthcare, environment, autonomous driving, and financial, requires reliable uncertainty estimates for individual predictions to avoid undesirable outcomes.
Uncertainty quantification (UQ) methods provide an approach to detect when predictions should not be trusted and when abstentions or human interventions are required.

The most commonly applied method for quantifying uncertainty is Bayesian model averaging (BMA) approximations, such as Monte Carlo dropout (MCD)~\cite{gal_dropout_2016}, deep ensembles (DE)~\cite{lakshminarayanan_simple_2017}, and more recently, last-layer ensembles (LLE)~\cite{gillis_last-layer_2025,schweighofer_introducing_2023,lee_why_2015}, by averaging or sampling from a posterior of finite model parameters (\hyperref[eq:bma]{Eq.~\ref{eq:bma}}). 
\begin{equation} \label{eq:bma}
    p(y \mid \mathbf{x}, \mathcal{D}) 
    \hspace{-0.05cm}
    =
    \hspace{-0.2cm}
    \int_{\mathbf{w}} p(y \mid \mathbf{x}, \mathbf{w}) \, p(\mathbf{w} \mid \mathcal{D}) \, \mathrm{dw}
    =
    \mathbb{E}_{\mathbf{w} \sim p(\mathbf{w} \mid \mathcal{D})} \left[ p(y \mid \mathbf{x}, \mathbf{w}) \right]
    \approx
    \frac{1}{M} \sum_{m=1}^{M} p(y \mid \mathbf{x}, \mathbf{w}_m)
\end{equation}
An important current debate is how to measure predictive uncertainty, and how this total uncertainty (TU) can be decomposed into aleatoric (AU) and epistemic (EU) uncertainty in order to evaluate if the uncertainty is inherent in the data or due to limitations of a model. A common approach is to use measures based on entropy (\hyperref[eq:bma]{Eq.~\ref{eq:population_mi}}).
\begin{equation} \label{eq:population_mi}
    \underbrace{
    \mathcal{H} \left[ p(y \mid \mathbf{x}, \mathcal{D}) \right]
    }_{\mathcal{TU} \text{ (Total Uncertainty)}}
    = \hspace{0.1cm}
    \underbrace{
    \mathbb{E}_{\mathbf{w} \sim p(\mathbf{w} \mid \mathcal{D})}
    \left[ 
    \mathcal{H}(p(y \mid \mathbf{x}, \mathbf{w}))
    \right] 
    }_{\mathcal{AU} \text{ (Aleatoric Uncertainty)}}
    ~~~+ 
    \underbrace{
    \mathcal{I} \left[ p(y, \mathbf{w} \mid \mathbf{x}, \mathcal{D}) \right]
    }_{\mathcal{EU} \text{ (Epistemic Uncertainty)}}
\end{equation}
Despite the theoretical support from information theory, the additive decomposition of total uncertainty has recently come under criticism and is now cautioned for uncertainty estimations~\cite{wimmer_quantifying_2023}.
Recent work by~\citet{schweighofer_introducing_2023} extensively examined the entropy-based approach and showed that it mistakenly assumed the BMA distribution to be equivalent to the true posterior predictive distribution. The authors subsequently introduced an expected pairwise cross-entropy and Kullback–Leibler (KL) divergence of ensemble model predictions as measures for predictive and epistemic uncertainty, which is arguably the current state-of-the-art method for uncertainty quantification.

In this study, we offer a complementary framework for measuring and decomposing uncertainty using the signal-to-noise-ratio (SNR) from means ($\mu$, signal) and standard deviations ($\sigma$, noise) of class probability distributions derived from from BMA of MCD, LLE (multihead) and MCD of LLE (multihead) ensembles.
We introduce a variance-gating function that attenuates class probabilities based on their local SNR.
After normalization, the rescaled distributions favors more confidently (\textit{i.e.}, reduced uncertainty) predicted regions of the probability space, along with a new predictive uncertainty profile. We also introduce a variance-gated measure of uncertainty and show preliminary data that our approach is comparable to \citet{schweighofer_introducing_2023}. Our experimental results revealed a challenge with committee machines in that after long training there is a collapse in the diversity of model samples. Although this has sometimes been briefly noted in the literature~\cite{schweighofer_introducing_2023,kirsch_implicit_2025,papyan_prevalence_2020,steger_function_2024}, we evaluated this phenomenon with our new measure. 

\section{Variance-Gated Predictive Distribution}
In the following, we consider a multiclass classification within a committee machine setting~\cite{gillis_last-layer_2025}.
To account for class variation in model disagreement, we introduce variance-gated distributions $\tilde{p}_{m,k}$ that combines the mean predicted probabilities $p(y \mid \mathbf{x}, \mathcal{D})$ with a per-class gating mechanism that is shared across ensemble models $m$ such that,
\begin{equation} \label{eq:gate}
    \tilde{p}_{m,k} =
    \frac{
        p_{m}{(y)} \cdot \Gamma_{k}(y)
    }{
        \sum\limits_{j} p_{m}{(j)} \cdot \Gamma_{k}(j)}
\end{equation}
where
$
\Gamma_{k}(y) = 
1 - \exp[-\mu{(y)}/{k\sigma(y)}]
$
and $\mu(y)$ denotes the ensemble mean predicted probability for class $y$ and $\sigma(y)$ is the corresponding ensemble standard deviation, which quantifies model-to-model disagreement.
The gating parameter $k > 0$ is a scalar hyperparameter that controls sensitivity to variance. The resulting gate $\Gamma_{k}(y)$ is a bounded, monotonic function in the range $[0, 1)$.
Under mild distributional assumptions, $k\sigma(y)$ may be viewed as a scale factor that reflects the typical deviation of ensemble members from the mean prediction, thus relating the gating function to the population of model divergences.
Effectively, hyperparameter $k$ allows a user to define how many ensemble members disagree or deviate from the mean.
This provides a used-defined sensitivity adjustment that can reflect the degree of acceptable risks for model predictions.
The resulting $\tilde{p}_{m,k}$ is subsequently normalized to ensure validity as a probability distribution. 
The variance-gated adjustment ensures that predictions with high model disagreement receive lower weight, even if their mean confidence is high.
When predictions are consistent (or collapsed) across the ensembles ($k\sigma(y) \ll \mu(y)$), we have $\Gamma_{k}(y) \rightarrow 1$ and thus $\tilde{p}_{m,k} \approx p$.
Conversely, when predictions are inconsistent ($k\sigma(y) \gg \mu(y)$); $\Gamma_{k}(y) \rightarrow 0$, so the correction reduces confidence such that $\tilde{p}_{m,k}(y) \rightarrow 0$.
As a result, uncertainty is reduced with increasing $k$-values since confident and certain predictions are preserved, while ambiguous and uncertain predictions are suppressed. A property that is lacking for the standard entropy-based decomposition measures~\cite{wimmer_quantifying_2023}.

We quantify predictive uncertainty using the variance-gated distribution $\tilde{p}_k(y \mid \mathbf{x}, \mathcal{D})$ with its entropy $\mathcal{H}[\tilde{p}_k]$ serving as the measure of total uncertainty. 
We obtain the principled decomposition of predictive uncertainty into aleatoric and epistemic components using variance-gated predictions defined as:  
\begin{equation} \label{eq:gated_decomposition}
    \mathcal{H} \left[ \tilde{p}_k(y \mid \mathbf{x}, \mathcal{D}) \right]
    = \hspace{0.1cm}
    \mathbb{E}_{\mathbf{w} \sim \tilde{p}_k(\mathbf{w} \mid \mathcal{D})}
    \left[ 
    \mathcal{H} (\tilde{p}_k(y \mid \mathbf{x}, \mathbf{w}))
    \right]
    +
    \mathcal{I} \left[ \tilde{p}_k(y, \mathbf{w} \mid \mathbf{x}, \mathcal{D}) \right]
\end{equation}
This framework provides a decomposition that respects ensemble variability and remains consistent with information-theoretic formulations (\hyperref[eq:population_mi]{Eq.~\ref{eq:population_mi}} \textit{vs.}~\hyperref[eq:gated_decomposition]{Eq.~\ref{eq:gated_decomposition}}). 
Using the standard uncertainty decomposition, epistemic uncertainty is obtained as the difference between total predictive uncertainty and the expected aleatoric uncertainty. 

\section{Variance-Gated Margin Uncertainty}  
The reliable estimation of uncertainty demands methods that goes beyond ensemble means.
Various strategies have met this requirement using entropy, including our proposed method that uses a combination of variance and entropy decomposition.
However, entropy is known to overestimate uncertainty when probability values are spread across many classes and underestimate when a model is highly confident between a few options.
Therefore, entropy alone is not sufficient to provide adequate information for decision-making.
We posit the following question: Can we identify a measure that is sensitive to class separation, incorporate uncertainty awareness, while avoiding the use of an entropy-based framework?
We want to identify a metric that 1) maintains epistemic awareness and 2) ideally, provides a user the ability to adjust risk-tolerance.

We propose to use the confidence margin of the top-2 mean predictions and corresponding standard deviations.
This approach is an extension of the Best-versus-Second Best (BvSB) introduced by~\citet{joshi_multi-class_2009}, where we incorporate variance along with a user-defined sensitivity hyperparameter $k$.  
Let $i$ and $j$ denote the top-1 and top-2 ranked classes by $\mu(\cdot),$ we define a prediction rule $\hat{y}$ and derive a SNR as:
\begin{equation}
    \text{SNR} = 
    \frac{\mu(i) - \mu(j)}
    {\sigma(i) + \sigma(j) + \epsilon} 
    > k,
    \qquad
    \hat{y} = 
    \begin{cases}
      i & \text{if}~\mu(i) - k\sigma(i) 
      >
      \mu(j) + k\sigma(j) \\
      \text{uncertain} & \text{otherwise}
    \end{cases}
\end{equation}
where $\mu{(i)} - \mu{(j)}$ is the probability margin between the two most likely classes and $\sigma{(i)} + \sigma{(j) + \epsilon}$ is the combined predictive variance, with $\epsilon > 0$ (\textit{e.g.}, $10^{-8}$) to ensures numerical stability.
In principle, the SNR can be interpreted as a binary decision boundary between classes $i$ and $j$, restricted by $k$. 
Under mild distribution assumptions, a user-defined threshold value of $k$ can be applied to reflect the fraction of samples requiring abstentions or human interventions.
For example, when $k=1$, only samples with $\text{SNR} > 1$ will be considered, all others are uncertain.
However, this criterion fails to capture cases where a model outputs ambiguous and uncertain predictions. 
Such outputs artificially inflates the SNR values, leading to misleading classifications.
To address this limitation, we introduce a variance-gated margin uncertainty (GMU), using a gating function that rescales model predictions by incorporating both confidence and variance (\textit{i.e.}, epistemic) information (\hyperref[eq:pcs]{Eq.~\ref{eq:pcs}}).
\begin{equation} \label{eq:pcs}
    \Gamma({i,j}) = 
    \left[1 - \exp\left(-\frac{
    \mu(i) - \mu(j)}{\sigma(i) + \sigma(j) + \epsilon
    } \right)\right],
    \qquad
    \text{GMU} = 1 - \mu(i) \cdot \Gamma(i,j)
    \qquad
\end{equation}
The GMU functions as a variance-gated uncertainty margin since large values corresponds to high separation of the top-1 and top-2 predictions, while small values capture situations of ambiguous and uncertain predictions.

\section{Experiments}
\label{sec:experiments}

\begin{figure}[t]
    \centering
    \begin{subfigure}{\linewidth}
        \centering
        \includegraphics[width=\linewidth]{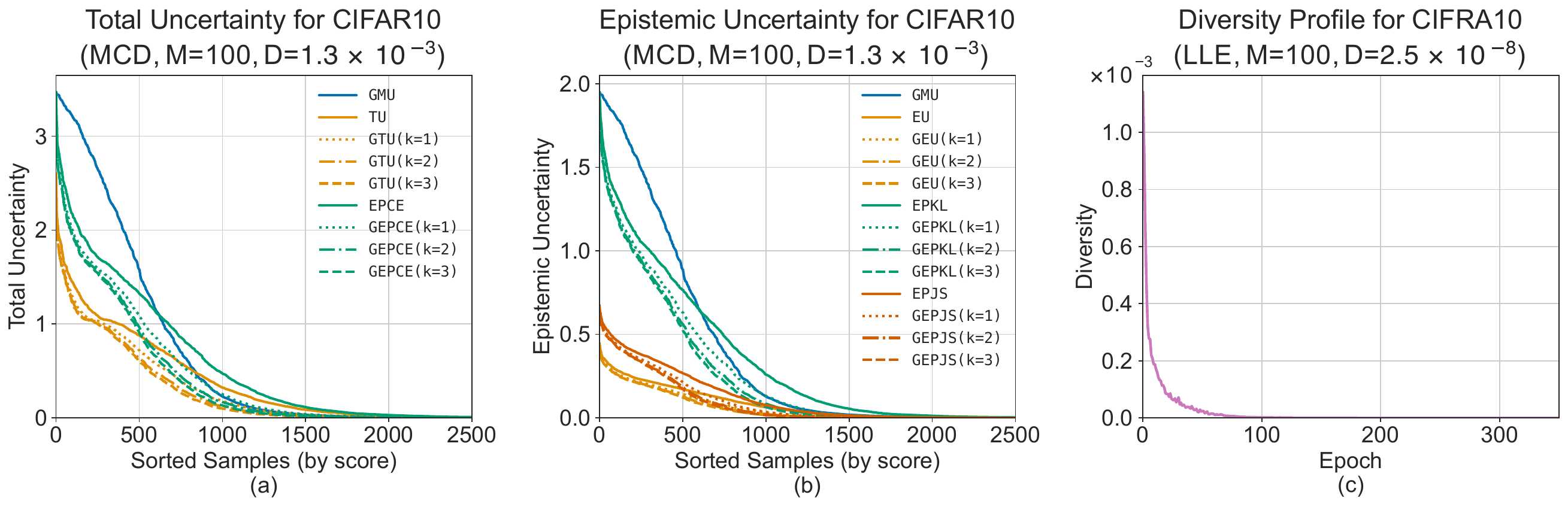}
    \end{subfigure}
    \begin{subfigure}{\linewidth}
        \centering
        \includegraphics[width=\linewidth]{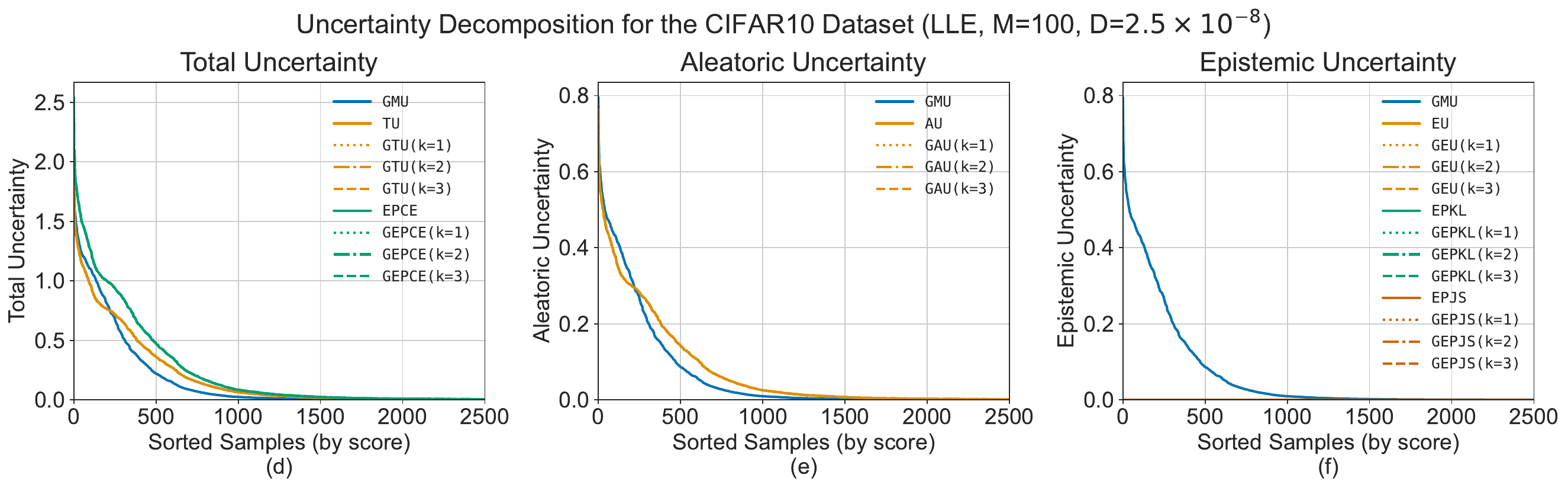}
    \end{subfigure}
    \caption{
    Uncertainty decomposition and diversity profile results for the CIFAR10 datasets using LLE and MCD ensembles.
    Panels (a, b, d--f) show the decomposition of total, aleatoric, and epistemic uncertainty using variance-gated distributions (GMU, GTU, GAU, GEU, GEPCE, GEPKL, GEPJS), compared against baseline measures (TU, AU, EU, EPCE, EPKL, EPJS).
    Panel (c) reports the diversity profile (LLE) across training epochs, quantified as the expectation over samples $i$ and classes $c$, of the variance across models ($\text{Diversity, D} = \mathbb{E}_{i,c}[\mathrm{Var}_M]$).
    Uncertainty measures were normalized (panels a, b, d, f) by the largest metric to allow direct comparison across methods.
    }
    \label{fig:lle_mcd_diversity}
\end{figure}

In the following, we present our preliminary results for the proposed variance-gated framework using standard benchmark datasets. 
Experiments were performed on MNIST, SVHN, CIFAR10, and CIFAR100, using three ensemble strategies: MCD, LLE, and MCD-LLE hybrids, each with 100 ensemble members. 
We compared our method against standard entropy-based uncertainty decomposition and the recent information-theoretic approach by~\citet{schweighofer_introducing_2023}, including:
1) total entropy (TU); 2) expected entropy (AU); 3) mutual information (EU); 4) expected pairwise cross-entropy (EPCE); 5) expected pairwise Kullback–Leibler divergence (EPKL); and 6) expected pairwise Jensen–Shannon divergence (EPJS).
As a representative example,~\hyperref[fig:lle_mcd_diversity]{Figure~\ref{fig:lle_mcd_diversity}} summarizes the decomposition of predictive uncertainty for the CIFAR10 dataset using LLE and MCD ensembles.
Additional experimental details and results for datasets, including a description for the multilabel GMU uncertainty measure, calibrations, and out-of-distribution (OOD) results for the CIFAR10 and SVHN datasets are provided in the~\hyperref[supp:supporting_information]{Supporting Information}.

\paragraph{Variance-Gated \textit{vs.}~Baseline Measures}
For the MCD ensembles (panels a and b), variance-gated uncertainty estimations and baseline measures agree on which samples are the most uncertain, approximately the first 1500 samples (out of 10000). 
Finer-grained or risk-adverse control can be obtained by selecting specific measures and/or adjusting the sensitivity hyperparameter $k$.
In this case, the variance-gated distributions produce decompositions that were consistently below non-gated estimations.
This confirms that the gating function attenuates predictions for high-variance, low-confidence predictions and estimations converge to baselines values with decreasing variance.
Our proposed GMU provided simar results, where the design of this measure is to identify ambiguous (\textit{i.e.}, top-2 predictions) cases where ensemble disagreement is high.
Therefore, it is quite possible that GMU results with a different set of uncertain samples. This is an active area we are currently investigating. 
Another important feature of GMU is its computational efficiency, since there are no nested calculations of entropy or pairwise divergences.

\paragraph{Diversity Collapse}

During our investigation, we observed that LLE networks had a tendency to collapse.
While this did not occur for all experiments, with extended training these networks converge to a single model.
We demonstrate this effect in panel (c) with the evolution of ensemble diversity over training epochs. 
While ensembles initially maintain high variability, diversity declines as training progresses, eventually collapsing to a near-zero variance state.
This indicates that despite large ensemble sizes (M=100), models can converge to similar solutions, reducing effective epistemic uncertainty.
Our proposed variance-gated distribution specifically highlights this observation where increasing $k$ has no effect on uncertainty estimations (panels d--f).
As a result, standard entropy-based and divergence-based methods may underestimate uncertainty once diversity collapses, since they implicitly assume disagreement across models.
Our variance-gated framework and GMU make this collapse explicit.
With decreasing diversity, gated estimations converge toward baseline predictions, signaling that the epistemic component has eroded. 

\section{Conclusions}
\label{sec:conclusions}
We proposed variance-gated distributions as a complementary approach for uncertainty decomposition and estimation.
Our results reveal a collapse of ensemble diversity during training, which variance-gated measures make explicit, providing a diagnostic tool for ensemble diversity and when diversity-preserving strategies may be required.
\newpage
\bibliography{_main}
%
\newpage
\appendix

\renewcommand{\thesection}{S\arabic{section}}
\renewcommand{\theequation}{S\arabic{equation}}
\renewcommand{\thefigure}{S\arabic{figure}}
\renewcommand{\thetable}{S\arabic{table}}
\renewcommand{\thepage}{S\arabic{page}}

\setcounter{section}{0}
\setcounter{equation}{0}
\setcounter{figure}{0}
\setcounter{table}{0}
\setcounter{page}{1}

\section*{\centering \huge Supporting Information}
\label{supp:supporting_information}

\section*{\LARGE Uncertainty Estimation using Variance-Gated Distributions}

\textbf{H. Martin Gillis, Isaac Xu, and Thomas Trappenberg}

\textit{Faculty of Computer Science, Dalhousie University, 6050 University Avenue, Halifax, NS  B3H 4R2, Canada}

\textit{E-mail: \href{mailto:tt@cs.dal.ca}{tt@cs.dal.ca} (Thomas Trappenberg)}

\vspace{1em}
\hrule height 1pt

\noindent \centering  \textbf{Table of Contents}
\vspace{-1em}

\startcontents[sections]
\printcontents[sections]{l}{1}{\setcounter{tocdepth}{2}}

\justifying
\vspace{1em}
\hrule height 0.5pt
\vspace{1pt}
\hrule height 0.5pt
%
\newpage
\section{Variance-Gated Predictive Uncertainty Decomposition}
\label{supp:vg_decomposition}

\subsection{Framework Definition and Setup}

We consider a multiclass classification setting where predictions are estimated from an ensemble model. For each input $\mathbf{x}$, we define (per-class):

\begin{equation} \label{eq:mu}
    \mu(y) = 
    \frac{1}{M} \sum_{m=1}^{M} p(y \mid \mathbf{x}, \mathbf{w}_m)
\end{equation}

\begin{equation} \label{eq:sigma}
    \sigma(y) = 
    \sqrt{\frac{1}{M} \sum_{m=1}^{M} \left( p(y \mid \mathbf{x}, \mathbf{w}_m) - \mu(y) \right)^2}
\end{equation}

where $\mu(y)$ is the ensemble mean prediction for class $y$ and $\sigma(y)$ is the corresponding predictive standard deviation.

We then define a \textbf{variance-gating function} as:

\begin{equation}
    \Gamma_{k}(y) = 
    \left[1 - \exp\left(-\frac{\mu(y)}{k\sigma(y) + \epsilon} \right)\right]
\end{equation}

where $k > 0$ is a tunable sensitivity hyperparameter and $\epsilon  > 0$ (\textit{e.g.}, $10^{-8}$) ensures numerical stability.

The \textbf{variance-gated normalized predictive distribution} is then defined as:
\begin{equation}
    \tilde{p}_{m,k} = \frac{p_m(y) \cdot \Gamma_{k}(y)}{\sum_j p_m(j) \cdot \Gamma_{k}(j)}
\end{equation}

\subsection{Uncertainty Decomposition}

Using Shannon entropy, we define the following:

\textbf{Total uncertainty (TU):}
\begin{equation}
    \mathcal{TU}_{\tilde{p}_k} = 
    \mathcal{H}[\tilde{p}_k] = -\sum_y \tilde{p}_{k}(y) \log \tilde{p}_{k}(y)
\end{equation}

Entropy of the variance-gated distribution quantifies total predictive uncertainty after variance-gated corrections.

\textbf{Aleatoric uncertainty (AU):}
\begin{equation}
    \mathcal{AU}_{\tilde{p}_k} =
    \mathbb{E}_{\mathbf{w} \sim \tilde{p}_k(\mathbf{w} \mid \mathcal{D})} \left[
    \mathcal{H}(\tilde{p}_k(y \mid \mathbf{x}, \mathbf{w}))
    \right]
    =
    \frac{1}{M} \sum_{m=1}^{M} \mathcal{H} \left[\tilde{p}_k(y \mid \mathbf{x}, \mathbf{w}_m)\right]  
\end{equation}
This is the aleatoric uncertainty that captures irreducible data noise and ambiguity. It is the standard formulation used by~\citet{houlsby_bayesian_2011,gal_dropout_2016,lakshminarayanan_simple_2017}~and~\citet{schweighofer_introducing_2023}.

\textbf{Epistemic uncertainty (EU):}
\begin{equation}
    \mathcal{EU}_{\tilde{p}_k} = 
    \mathcal{TU}_{\tilde{p}_k} - \mathcal{AU}_{\tilde{p}_k}
\end{equation}

This is the residual model uncertainty after removing ensemble variability with the variance-gated distribution.
It captures the reduction of confidence due to ensemble disagreement (\textit{i.e.}, epistemic).

\subsection{Properties}

\paragraph{Normalization}
The gating function ensures that $\tilde{p}_{m,k}$ is a valid probability distribution given,
$p_m(y) \cdot \Gamma_{k}(y) \geq 0 \quad \text{and} \quad \sum_y \tilde{p}_{m,k}(y) = 1$.

\paragraph{Limit Behavior}

\begin{align*}
    & 
    \lim_{\substack{\mu(y) \to 1 \\ k\sigma(y) \to 0^+}} 
    \hspace{0.2cm}
    p_m(y) \times \left[1 - \exp\left(-\frac{1}{\epsilon}\right)\right] 
    \approx
    p_m(y) \quad \quad \text{(confident, certain)}
    \\
    &
    \lim_{\substack{\mu(y) \to 1 \\ k\sigma(y) \to \infty}} 
    \hspace{0.2cm}
    p_m(y) \times \left[1 - \exp\left(-\frac{1}{\infty}\right)\right] 
    =
    0 \quad \quad \text{(confident, uncertain)}
    \\
    & 
    \lim_{\substack{\mu(y) \to 0^+ \\ k\sigma(y) \to 0^+}} 
    \hspace{0.2cm}
    p_m(y) \times \left[1 - \exp\left(-\frac{0^+}{\epsilon}\right)\right] 
    = 
    0 \quad \quad \quad \text{(ambiguous, certain)}
    \\
    & 
    \lim_{\substack{\mu(y) \to 0^+ \\ k\sigma(y) \to \infty}} 
    \hspace{0.2cm}
    p_m(y) \times \left[1 - \exp\left(-\frac{0^+}{\infty}\right)\right] 
    =
    0 \quad \quad \quad \text{(ambiguous, uncertain)}
\end{align*}

\paragraph{Boundedness}
For all $k > 0$, $\mu(y) \in [0, 1]$, and $\sigma(y) \geq 0$, the gating function is bounded such that,
$0 \leq p_m(y) \cdot \Gamma_{k}(y) < p_m(y)$.
This bound ensures that uncertainty suppresses confidence without exceeding ensemble agreement. 

\paragraph{Distributional Assumption}  
The variance-gated formulation assumes an approximate normal distribution of class probabilities across model predictions.
However, other distributional models can be employed and implemented (\textit{e.g.}, Gaussian mixture models).

\subsection{Summary of Equations}

\begin{equation*}
    \boxed{
    \begin{alignedat}{2}
    \Gamma_{k}(y) 
    &= 
    \left[1 - \exp\!\left(-\frac{\mu(y)}{k\sigma(y) + \epsilon} \right)\right] 
    & \qquad 
    \mathcal{TU}_{\tilde{p}_k} 
    &= 
    -\sum_y \tilde{p}_{k}(y) \log \tilde{p}_{k}(y)
    \\
    \tilde{p}_{m,k} 
    &= 
    \frac{p_m(y)\cdot \Gamma_{k}(y)}{\sum_j p_m(j) \cdot \Gamma_{k}(j)} 
    &\qquad 
    \mathcal{AU}_{\tilde{p}_k} 
    &= 
    \frac{1}{M} \sum_{m=1}^{M} \mathcal{H} \left[\tilde{p}_k(y \mid \mathbf{x}, \mathbf{w}_m)\right] 
    \\
    & &\qquad 
    \mathcal{EU}_{\tilde{p}_k} 
    &= 
    \mathcal{TU}_{\tilde{p}_k} - \mathcal{AU}_{\tilde{p}_k}
    \end{alignedat}
    }
\end{equation*}

\section{Variance-Gated Margin Uncertainty: Multiclass}
\label{supp:mc_pcs}

\subsection{Framework Definition and Setup} 

As discussed in the main text.

\subsection{Properties}

\paragraph{Limit Behavior}
We analyze the behavior under four distinct limit regimes, defined by whether the mean difference tends to 1 or 0 and whether the uncertainty tends to $\infty$ or 0.

\begin{align*}
    & 
    \lim_{\substack{\mu(i) - \mu(j) \to 1 \\ \sigma(i) + \sigma(j) \to 0^+}} 
    \hspace{0.2cm}
    \mu(i) \times \left[1 - \exp\left(-\frac{1}{\epsilon}\right)\right] 
    \approx 
    \mu(i) \quad \quad \text{(confident, certain)}
    \\
    &
    \lim_{\substack{\mu(i) - \mu(j) \to 1 \\ \sigma(i) + \sigma(j) \to \infty}} 
    \hspace{0.2cm}
    \mu(i) \times \left[1 - \exp\left(-\frac{1}{\infty}\right)\right] 
    =
    0 \quad \quad \text{(confident, uncertain)}
    \\
    & 
    \lim_{\substack{\mu(i) - \mu(j) \to 0^+ \\ \sigma(i) + \sigma(j) \to 0^+}} 
    \hspace{0.2cm}
    \mu(i) \times \left[1 - \exp\left(-\frac{0^+}{\epsilon}\right)\right] 
    =
    0 \quad \quad \quad \text{(ambiguous, certain)}
    \\
    & 
    \lim_{\substack{\mu(i) - \mu(j) \to 0^+ \\ \sigma(i) + \sigma(j) \to \infty}} 
    \hspace{0.2cm}
    \mu(i) \times \left[1 - \exp\left(-\frac{0^+}{\infty}\right)\right] 
    = 
    0 \quad \quad \quad \text{(ambiguous, uncertain)}
\end{align*}

\paragraph{Boundedness}
For all $\mu(i) - \mu(j) \in [0, 1]$ and $\sigma(i) + \sigma(j) \geq 0$, 
the gating function $\Gamma(i,j)$ satisfies

\begin{equation*}
    0 \leq \Gamma(i,j) < 1
    \quad \implies \quad
    0 \leq \mu(i) \cdot \Gamma(i,j) < \mu(i)
\end{equation*}

This guarantees that the confidence gate only reduces the predicted score relative to the ensemble model raw agreements $\mu(i)$.

\paragraph{Distributional Assumption}  
As noted for the variance-gated normalized predictive distribution, the GMU formulation assumes an approximate normal distribution.
Here again, alternative distributional models can be validated and implemented with the framework.

\subsection{Summary of Equations}

\begin{equation*}
    \boxed{
    \begin{aligned}
        \hat{y} 
        &=
        \begin{cases}
        i & \text{if}~\mu(i)-k\sigma(i) > \mu(j)+k\sigma(j) \\ 
        \text{uncertain} & \text{otherwise}
        \end{cases}
        & 
        \\
        \text{SNR} 
        &= \frac{\mu(i)-\mu(j)}{\sigma(i)+\sigma(j)+\epsilon} > k
        & 
        \\
        \Gamma(i,j) &= 
        \left[
        1 - \exp\!\left(-\frac{\mu(i)-\mu(j)}{\sigma(i)+\sigma(j)+\epsilon}\right)
        \right]
        \\
        \text{GMU} 
        &= 1 - \mu(i) \cdot \Gamma(i,j)
    \end{aligned}
    }
\end{equation*}

\section{Variance-Gated Margin Uncertainty: Multilabel} 
\label{supp:ml_pcs}

\subsection{Framework Definition and Setup}

We consider the a multilabel classification setting, where predictions are estimated from an ensemble model (\hyperref[eq:mu]{Eq.~\ref{eq:mu}} and~\hyperref[eq:sigma]{Eq.~\ref{eq:sigma}}) and derive a variance-gating function based on the signal-to-noise ratio (SNR) of the top-2 predictions.
In this case, the top-1 prediction is defined as 
$\mu(i) = \max(u(i), 1 - u(i))$
and the corresponding top-2 prediction is the complementary value 
$\mu(j) = 1 - \mu(i)$, such the $\mu(i) + \mu(j) = 1$ and $\sigma(i) = \sigma(j)$.

We define a prediction rule and derive a SNR as:

\begin{equation}
    \text{SNR} = 
    \frac{2\mu(i) - 1}
    {2\sigma(i) + \epsilon} 
    > k,
    \qquad
    \hat{y} = 
    \begin{cases}
      i & \text{if}~\mu(i) - k\sigma(i) 
      >
      (1 - \mu(i)) + k\sigma(i) \\
      \text{uncertain} & \text{otherwise}
    \end{cases}
\end{equation}

where $2\mu(i)- 1$ is the probability margin between the top-1 prediction with its complement and $2\sigma(i) + \epsilon$ is the combined predictive spread, with $\epsilon > 0$ (\textit{e.g.}, $10^{-8}$) to ensures numerical stability.
This is exactly analogous to the multiclass setting; however, the top-2 predictions are derived from a single multilabel prediction.

For a multilabel setting, the SNR can be interpreted as a binary decision boundary for label $i$ being present or not, restricted by $k$. 
We then define a \textbf{multilabel variance-gated margin uncertainty (GMU)}, using a variance-gating function that rescales model predictions by incorporating both confidence and variance information (as done previously).

\begin{equation}
    \text{GMU} 
    =
    \begin{cases}
    1- \mu(i) \cdot \Gamma({i}) & \text{if}~\mu(i) > 0.5 \\ 
    \mu(i) \cdot \Gamma({i}) & \text{otherwise}
    \end{cases},
    \qquad
    \Gamma({i}) = 
    \left[1 - \exp\left(-\frac{
    2\mu(i) - 1}{2\sigma(i) + \epsilon
    } \right)\right]
\end{equation}

\subsection{Properties}

\paragraph{Limit Behavior}

\begin{align*}
    & 
    \lim_{\substack{\mu(i) \to 1 \\ \sigma(i) \to 0^+}} 
    \hspace{0.2cm}
    \mu(i) \times \left[1 - \exp\left(-\frac{1}{\epsilon}\right)\right] 
    \approx 
    \mu(i) \quad \quad \text{(confident, certain)}
    \\
    &
    \lim_{\substack{\mu(i) \to 1 \\ \sigma(i) \to \infty}} 
    \hspace{0.2cm}
    \mu(i) \times \left[1 - \exp\left(-\frac{1}{\infty}\right)\right] 
    =
    0 \quad \quad \text{(confident, uncertain)}
    \\
    & 
    \lim_{\substack{\mu(i) \to 0^+ \\ \sigma(i) \to 0^+}} 
    \hspace{0.2cm}
    \mu(i) \times \left[1 - \exp\left(-\frac{0^+}{\epsilon}\right)\right] 
    =
    0 \quad \quad \quad \text{(ambiguous, certain)}
    \\
    & 
    \lim_{\substack{\mu(i) \to 0^+ \\ \sigma(i) \to \infty}} 
    \hspace{0.2cm}
    \mu(i) \times \left[1 - \exp\left(-\frac{0^+}{\infty}\right)\right] 
    = 
    0 \quad \quad \quad \text{(ambiguous, uncertain)}
\end{align*}

\paragraph{Boundedness}
For all $2\mu(i) - 1 \in [0, 1]$ and $2\sigma(i) \geq 0$, 
the gating function $\Gamma(i)$ satisfies:

\begin{equation*}
    0 \leq \Gamma(i)< 1
    \quad \implies \quad
    0 \leq \mu(i) \cdot \Gamma(i)~\text{or}~(1-\mu(i)) \cdot \Gamma(i) < \mu(i)
\end{equation*}

This guarantees that the confidence gate only reduces the predicted score relative to the ensemble model raw agreements $\mu(i)$.

\paragraph{Distributional Assumption}  
Same as multiclass setting.

\subsection{Summary of Equations}

\begin{equation*}
    \boxed{
    \begin{aligned}
        \hat{y} 
        &=
        \begin{cases}
        i & \text{if}~\mu(i)-k\sigma(i) > (1 - \mu(i)) + k\sigma(i) \\ 
        \text{uncertain} & \text{otherwise}
        \end{cases}
        & 
        \\
        \text{SNR} 
        &= \frac{2\mu(i) - 1}{2\sigma(i) + \epsilon} > k
        & 
        \\
        \Gamma(i) 
        &= 
        \left[
        1 - \exp\!\left(-\frac{2\mu(i) - 1}{2\sigma(i) + \epsilon}\right)
        \right]
        \\
        \text{GMU} 
        &= 
        \begin{cases}
        1 - \mu(i) \cdot \Gamma({i}) & \text{if}~\mu(i) > 0.5 \\ 
        \mu(i) \cdot \Gamma({i}) & \text{otherwise}
        \end{cases}
    \end{aligned}
    }
\end{equation*}

\section{Experimental Details}
\label{supp:experimental_details}

\subsection{Datasets}
The MNIST, SVHN, CIFAR10/100 datasets were obtained from the PyTorch dataset repository.
Dataset means and standard deviations were computed and used during training and inference.
For OOD experiments, OOD samples were normalized using the in-domain (ID) means and standard deviations.

\subsection{Network Architectures, Training, and Calibration}
\paragraph{MNIST}
For the MNIST dataset, a CNN was used for MCD, LLE, and MCD-LLE networks.
Here, the CNN consisted of a feature extraction module with two blocks of convolutions, ReLU activations, maxpooling, and a dropout layer ($p=0.05$).
Both blocks used convolution with 128 output channels, a kernel size of 5, and a maxpooling layer with a kernel and stride size of 2.
The second block was subsequently flattened before being passed to a linear layer (2048 input and 2048 output channels), ReLU activation, and a dropout layer ($p=0.05$).
A classifier was added which consisted of a linear layer (2048 input and 10 output channels).
In the case of the LLE, the classifier was converted to a list of 100 classifiers.
Training was done with a batch size of 128 over 100 epochs using the Adam optimizer with a learning rate of $1.0 \times 10^{-5}$ and the cross-entropy loss objective function (for each classifier).
Mean classifier loss was then calculated before being backpropagated during optimization.
Calibration was performed by finding the optimal temperature (\textit{i.e.}, scaling of logits) that minimizes the cross-entropy loss on the validation dataset. 
This was achieved using a pre-trained model and Bayesian optimization with Gaussian Process (GP) regression~\cite{louppe_bayesian_2016}. 
The optimization explored a temperature search space of 0.01--10.0, using a log-uniform prior which was configured to perform 50 evaluations of the objective function.
For LLE, the temperature for each committee member was optimized separately.
After finding the optimal temperature, the pre-trained model was used to perform inference on the testing dataset, scaling the logits with the optimized calibrated temperature(s).

\paragraph{SVHN, CIFAR10, and CIFAR100}
The WideResNet-28-10 network architecture was used as described by~\citet{zagoruyko_wide_2017} with a dropout rate of $p=0.05$ for SVHN and $p=0.3$ for CIFAR10/100 datasets.
For LLE, a list of classifiers was added as described for MNIST dataset.
The SVHN dataset was trained over 100 epochs with a batch size of 128 using the Adam optimizer with a learning rate of $1.0 \times 10^{-6}$ and the cross-entropy loss objective function.
A cosine annealing scheduler was applied during training using an initial and final learning rate factor of 0.1, where the rate was increased over the first 25\% of the total number of epochs. 
CIFAR10 and CIFAR100 were trained over 250 or 350 epochs with a batch size of 128 using the Adam optimizer with a learning rate of $1.0 \times 10^{-4}$ and
a weight decay of $1.0 \times 10^{-5}$. The same learning rate scheduling strategy was applied and for the SVHN dataset.
Network calibrations for SVHN, CIFAR10, and CIFAR100 were done as described for the MNIST dataset.

\section{Additional Experimental Results}
\label{supp:additional_experimental_results}

This section presents supplementary experimental results across multiple benchmark datasets, providing a broader validation of our methods. 
We provide an overview of performance in~\hyperref[tab:summary_results]{Table~\ref{tab:summary_results}}, followed by analyses for the MNIST, SVHN, CIFAR10, and CIFAR100 datasets. 
For each dataset, we include results using different ensemble and calibration settings, such as MCD, MCD-LLE, LLE, and LLE (calibrated). 
The figures illustrate how the proposed framework behaves across increasingly complex tasks, identifying collapse, and the impact of calibration on predictive uncertainty estimates. 
These results demonstrate that the variance-gated approach generalizes across diverse data and supports the main claims of the paper.
\begin{table}[b]
    \centering
    \begin{threeparttable}
        \caption{Performance and calibration metrics for ensemble models.}
        \begin{tabular}{ l l l l l }
            \toprule
            \textbf{Dataset}
            & \textbf{Ensemble}
            & \textbf{Accuracy}
            & \textbf{F1-score}
            & \textbf{ECE}\tnote{1}
            \\
            \midrule
            MNIST & 
            MCD & 
            0.992 &
            0.992 &
            0.002 \\
            & 
            MCD-LLE & 
            0.994 &
            0.994 &
            0.003 \\
            & 
            LLE & 
            0.994 &
            0.994 &
            0.002 \\
            & 
            LLE (calibrated) & 
            0.994 &
            0.994 &
            0.002 \\
            \midrule
            SVHN & 
            MCD & 
            0.920 &
            0.913 &
            0.030 \\
            & 
            MCD-LLE & 
            0.924 &
            0.916 &
            0.051 \\
            & 
            LLE & 
            0.992 &
            0.911 &
            0.019 \\
            & 
            LLE (calibrated) & 
            0.921 &
            0.911 &
            0.007 \\
            \midrule
            CIFAR10 & 
            MCD & 
            0.956 &
            0.956 &
            0.017 \\
            & 
            MCD-LLE & 
            0.957 &
            0.957 &
            0.014 \\
            & 
            LLE & 
            0.957 &
            0.957 &
            0.025 \\
            & 
            LLE (calibrated) & 
            0.957 &
            0.957 &
            0.006 \\
            \midrule
            CIFAR100 & 
            MCD & 
            0.771 &
            0.771 &
            0.075 \\
            & 
            MCD-LLE & 
            0.775 &
            0.775 &
            0.049 \\
            & 
            LLE & 
            0.772 &
            0.770 &
            0.105 \\
            & 
            LLE (calibrated) & 
            0.772 &
            0.770 &
            0.025 \\
            \midrule
            \bottomrule
        \end{tabular}
        $^1$ Expected calibration error.
        \label{tab:summary_results} 
    \end{threeparttable}
\end{table}

\subsection{MNIST}

\begin{figure}[!ht]
    \centering
    \begin{subfigure}{\linewidth}
        \centering
        \includegraphics[width=\linewidth]{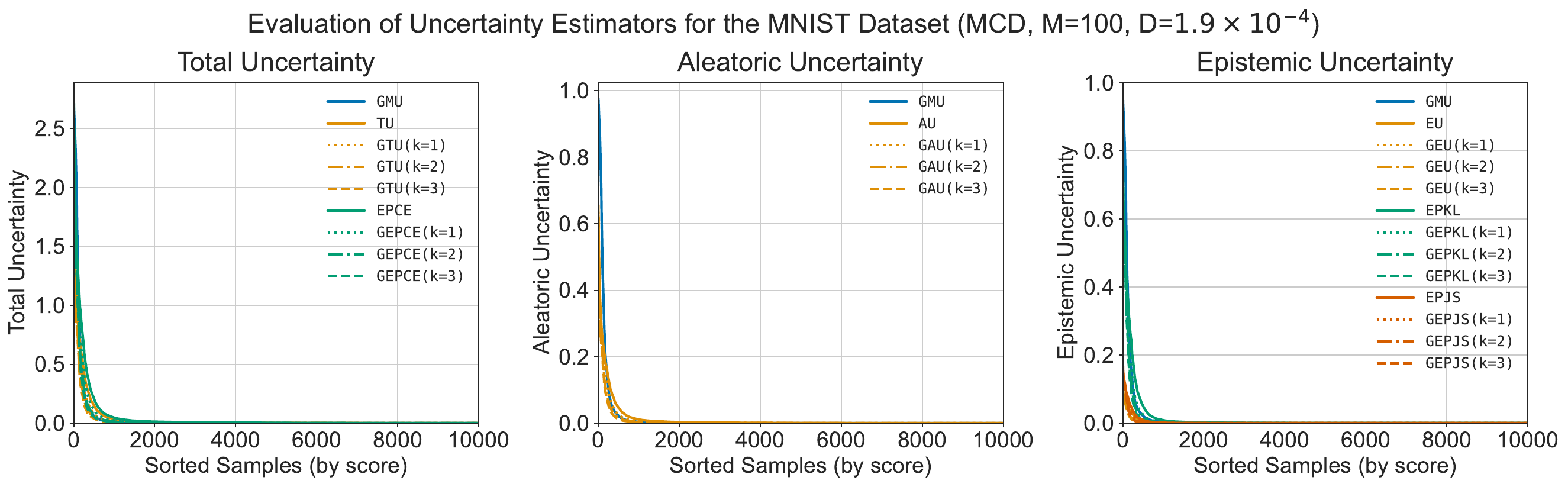}
    \end{subfigure}
    \begin{subfigure}{\linewidth}
        \centering
        \includegraphics[width=\linewidth]{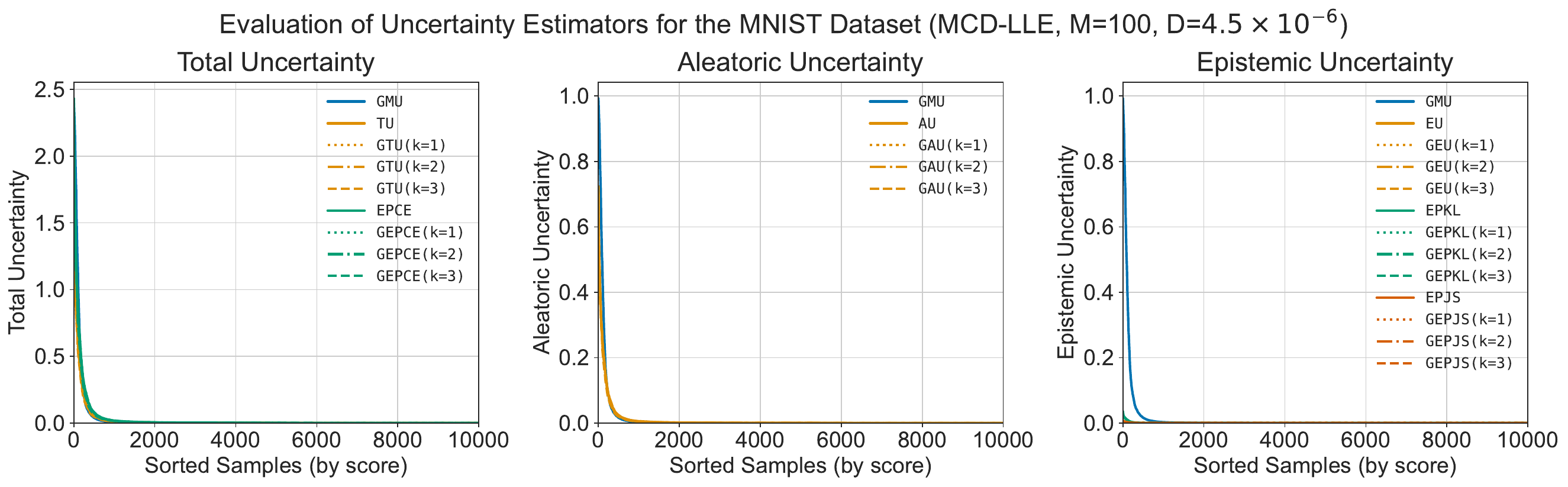}
    \end{subfigure}
    \begin{subfigure}{\linewidth}
        \centering
        \includegraphics[width=\linewidth]{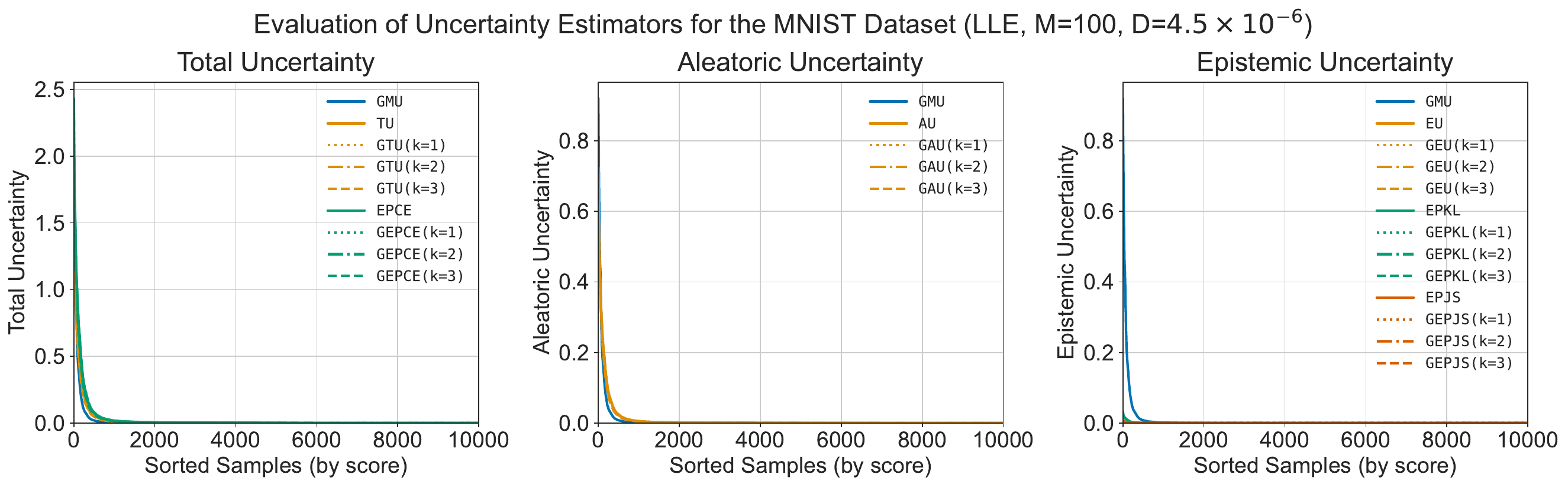}
    \end{subfigure}
    \begin{subfigure}{\linewidth}
        \centering
        \includegraphics[width=\linewidth]{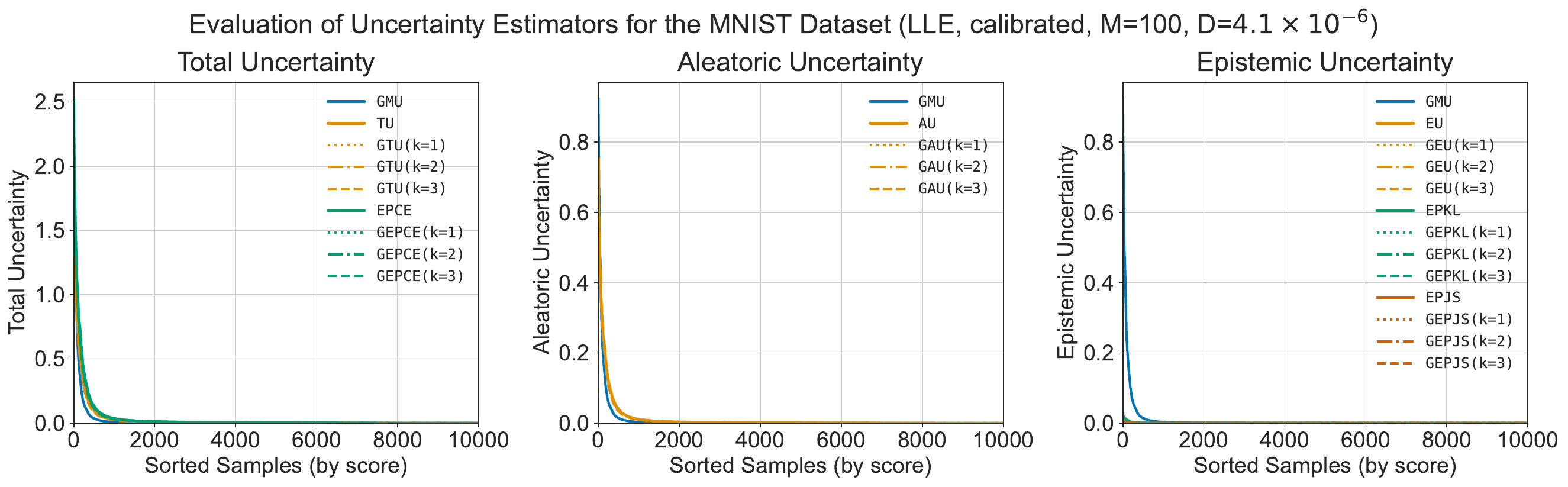}
    \end{subfigure}
    \caption{
    Uncertainty decompositions for the \textbf{MNIST} dataset using \textbf{MCD, MCD-LLE, LLE, and LLE (calibrated)} networks and variance-gated distributions, compared against baseline measures.
    Diversity (D) was quantified as the expectation over samples $i$ and classes $c$, of the variance across models ($\mathbb{E}_{i,c}[\mathrm{Var}_M]$). 
    Uncertainty measures were normalized by the largest metric to allow direct comparison across methods.
    }
\end{figure}

\clearpage

\subsection{SVHN}

\begin{figure}[!ht]
    \centering
    \begin{subfigure}{\linewidth}
        \centering
        \includegraphics[width=\linewidth]{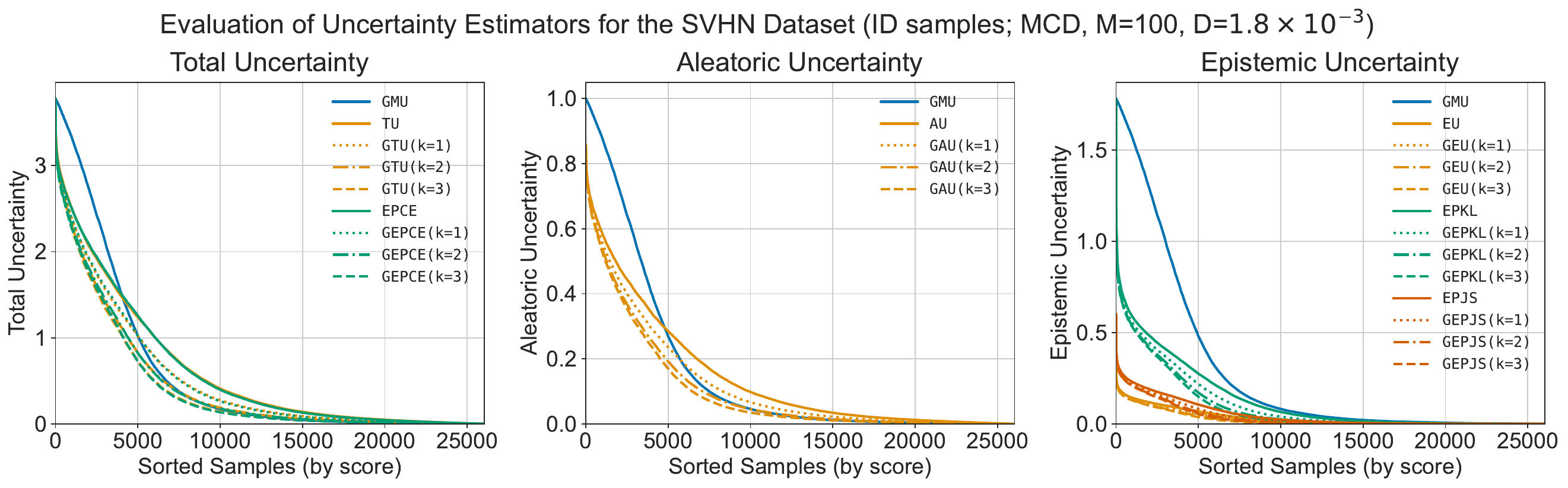}
    \end{subfigure}
    \begin{subfigure}{\linewidth}
        \centering
        \includegraphics[width=\linewidth]{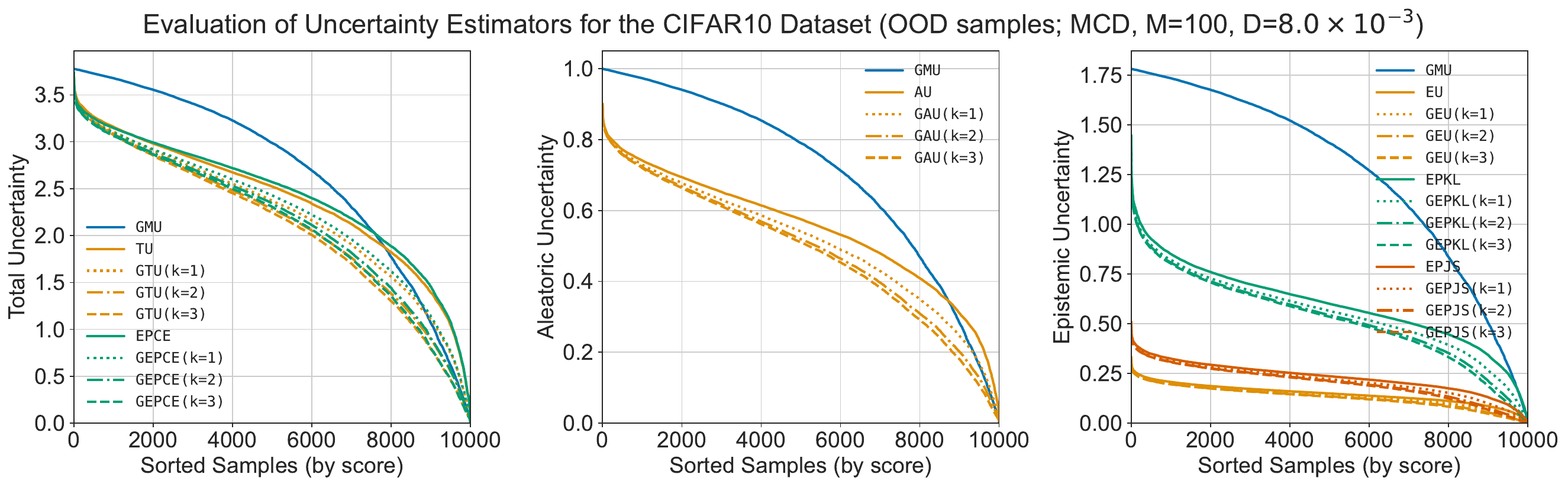}
    \end{subfigure}
    \begin{subfigure}{\linewidth}
        \centering
        \includegraphics[width=\linewidth]{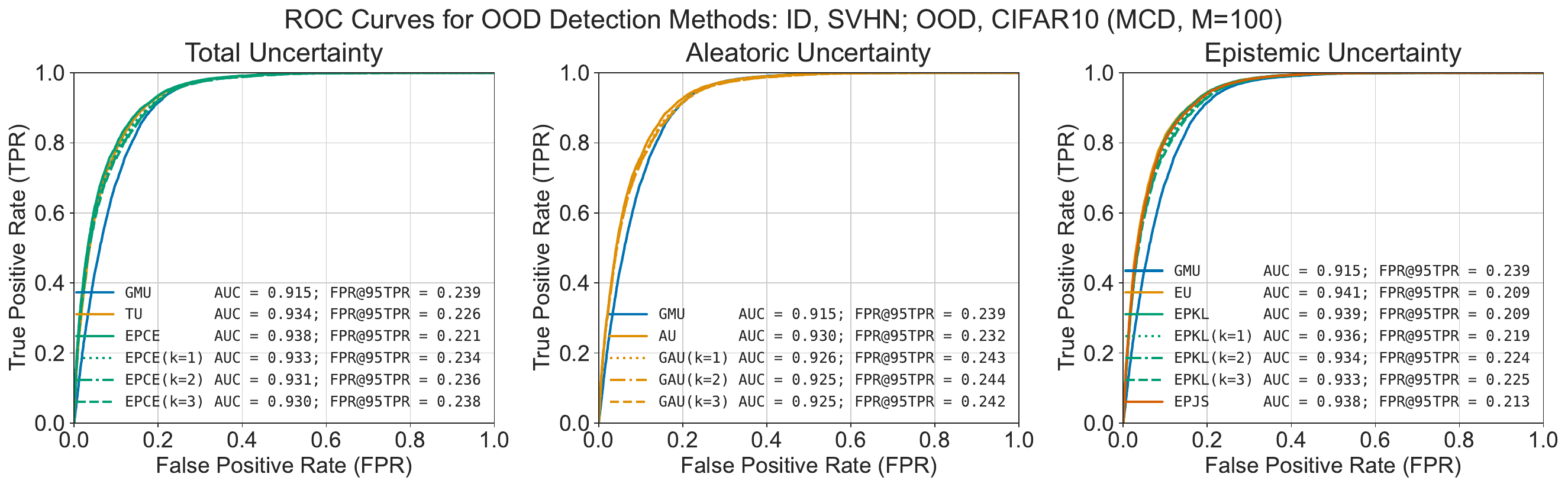}
    \end{subfigure}
    \caption{
    \textbf{Uncertainty decompositions} and \textbf{OOD} for the \textbf{SVHN} dataset using \textbf{MCD} network and variance-gated distributions, compared against baseline measures.
    Diversity (D) was quantified as the expectation over samples $i$ and classes $c$, of the variance across models ($\mathbb{E}_{i,c}[\mathrm{Var}_M]$). 
    Uncertainty measures were normalized by the largest metric to allow direct comparison across methods.
    }
\end{figure}

\clearpage

\begin{figure}[!ht]
    \centering
    \begin{subfigure}{\linewidth}
        \centering
        \includegraphics[width=\linewidth]{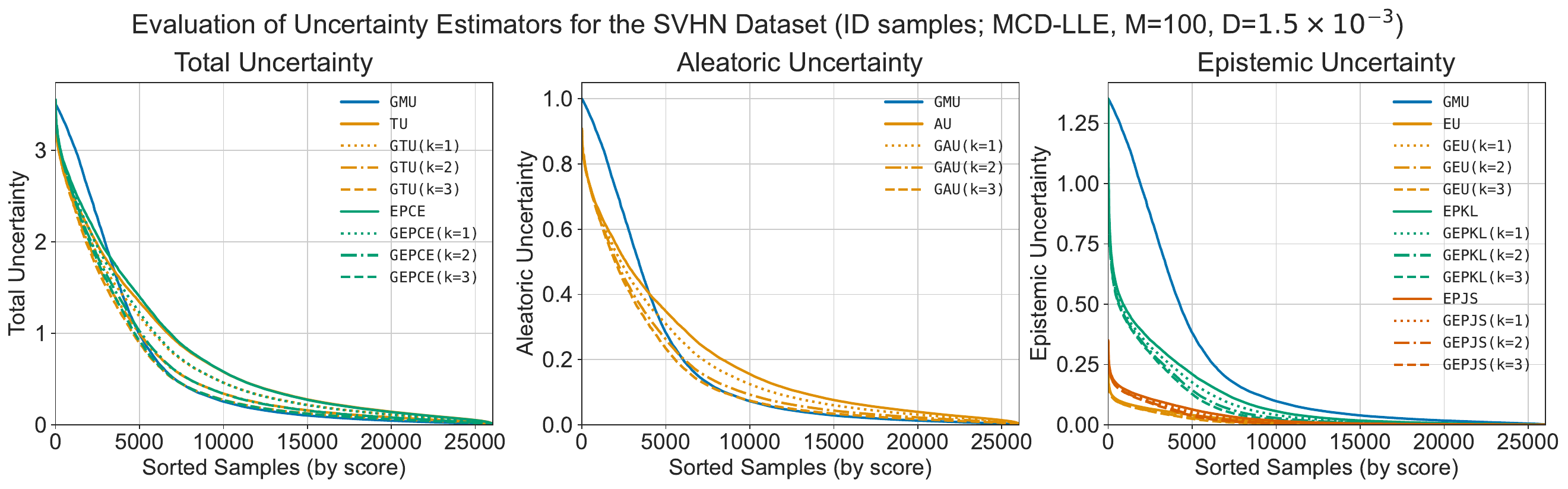}
    \end{subfigure}
    \begin{subfigure}{\linewidth}
        \centering
        \includegraphics[width=\linewidth]{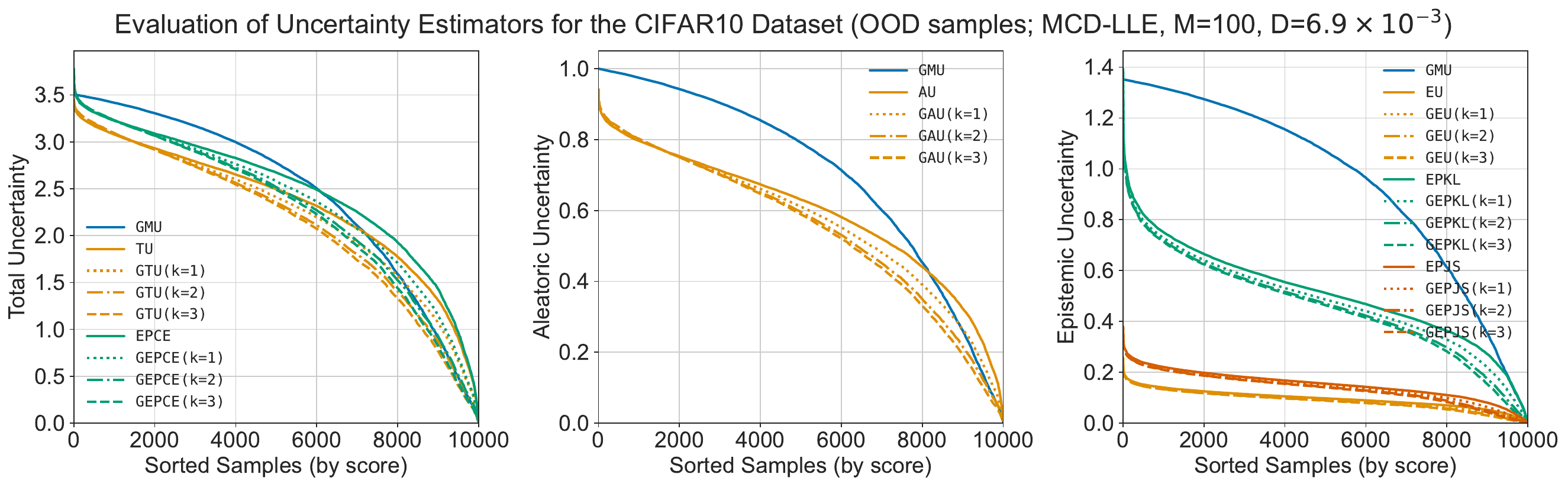}
    \end{subfigure}
    \begin{subfigure}{\linewidth}
        \centering
        \includegraphics[width=\linewidth]{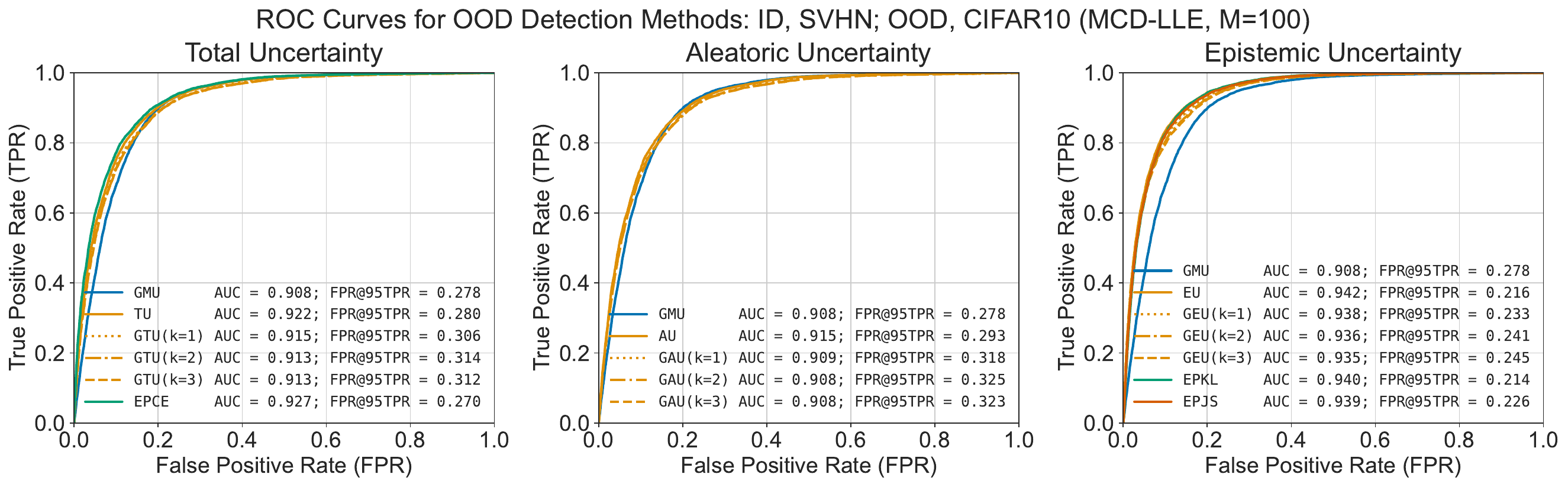}
    \end{subfigure}
    \caption{
    \textbf{Uncertainty decompositions} and \textbf{OOD} for the \textbf{SVHN} dataset using \textbf{MCD-LLE} network and variance-gated distributions, compared against baseline measures.
    Diversity (D) was quantified as the expectation over samples $i$ and classes $c$, of the variance across models ($\mathbb{E}_{i,c}[\mathrm{Var}_M]$). 
    Uncertainty measures were normalized by the largest metric to allow direct comparison across methods.
    }
\end{figure}

\clearpage

\begin{figure}[!ht]
    \centering
    \begin{subfigure}{\linewidth}
        \centering
        \includegraphics[width=\linewidth]{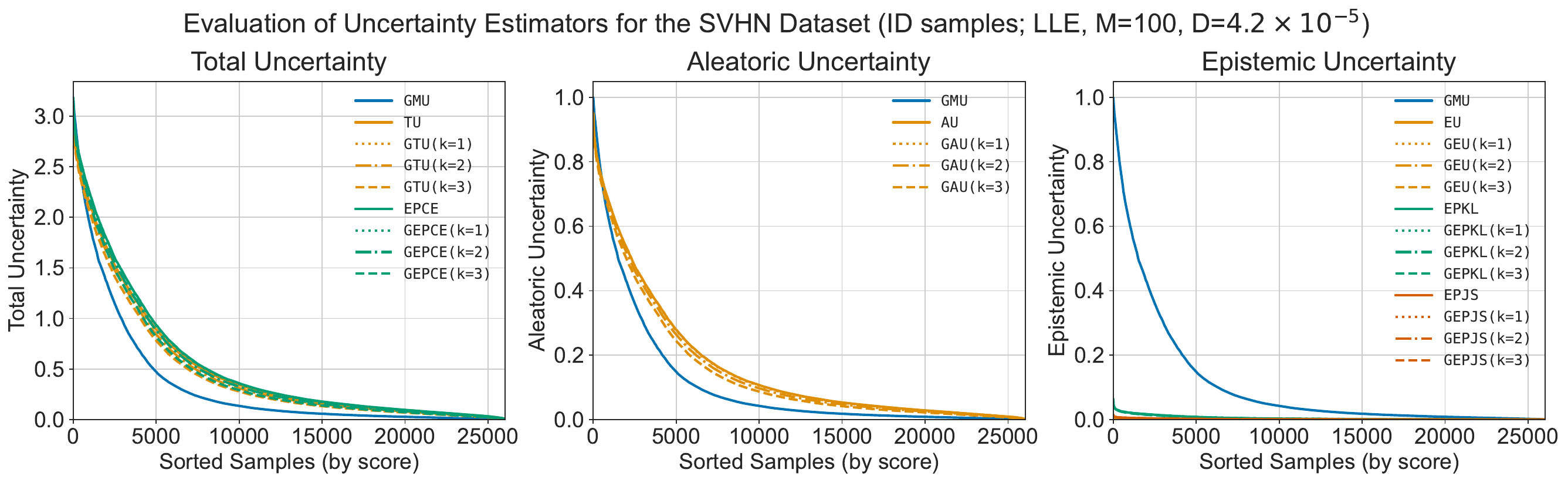}
    \end{subfigure}
    \begin{subfigure}{\linewidth}
        \centering
        \includegraphics[width=\linewidth]{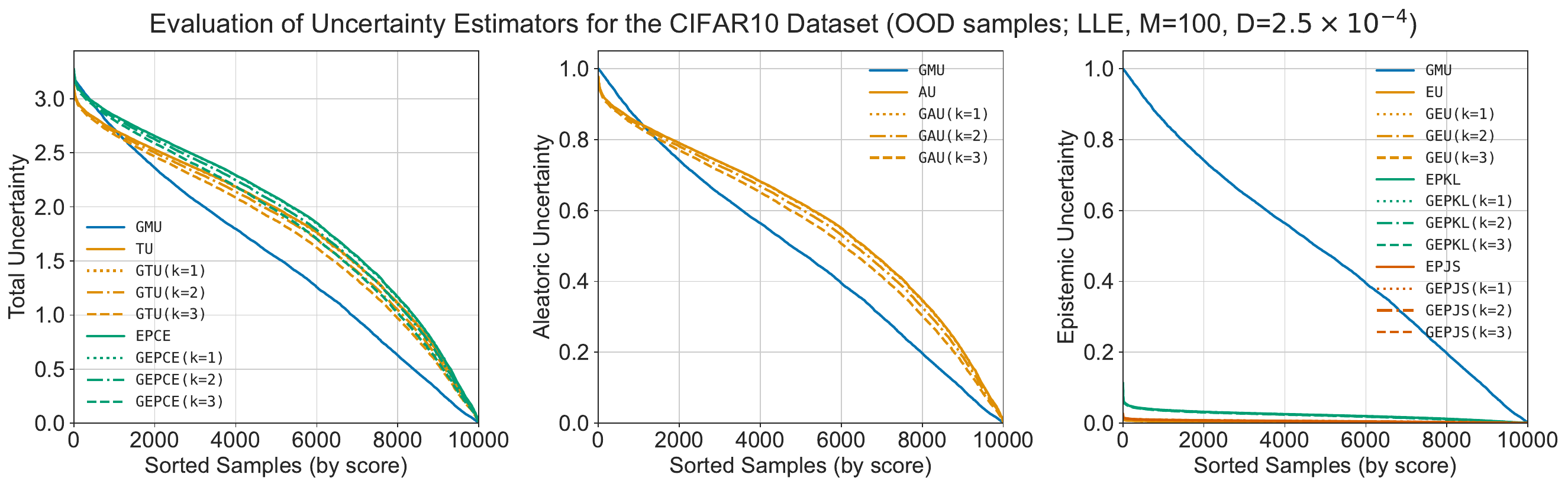}
    \end{subfigure}
    \begin{subfigure}{\linewidth}
        \centering
        \includegraphics[width=\linewidth]{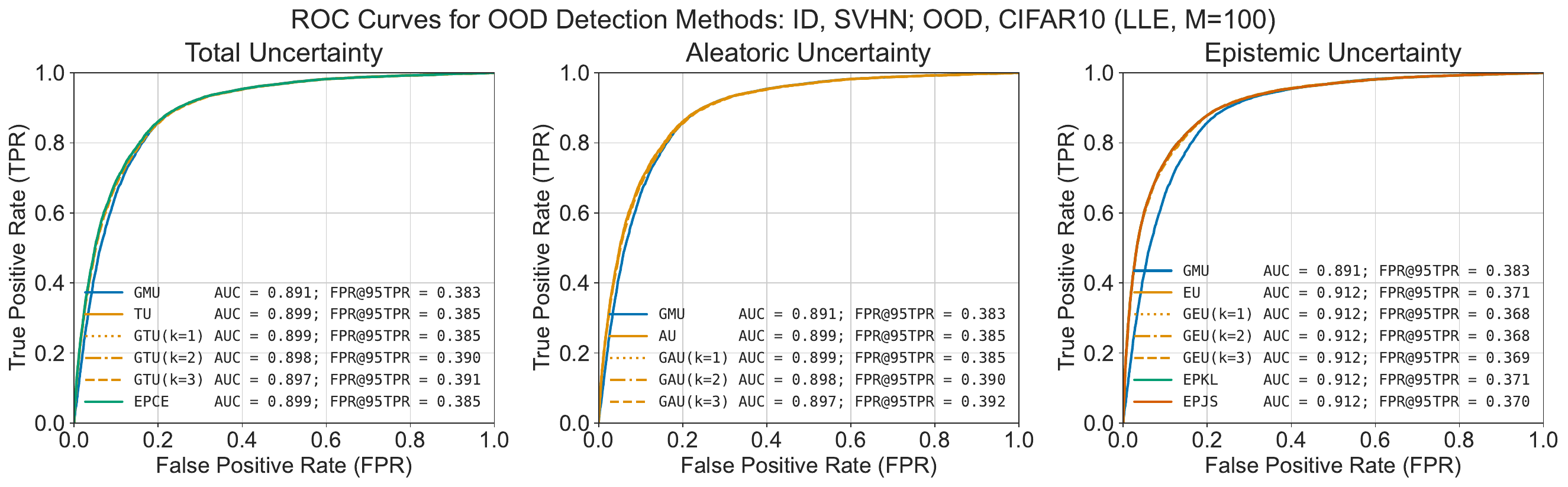}
    \end{subfigure}
    \caption{
    \textbf{Uncertainty decompositions} and \textbf{OOD} for the \textbf{SVHN} dataset using \textbf{LLE} network and variance-gated distributions, compared against baseline measures.
    Diversity (D) was quantified as the expectation over samples $i$ and classes $c$, of the variance across models ($\mathbb{E}_{i,c}[\mathrm{Var}_M]$). 
    Uncertainty measures were normalized by the largest metric to allow direct comparison across methods.
    }
\end{figure}

\clearpage

\begin{figure}[!ht]
    \centering
    \begin{subfigure}{\linewidth}
        \centering
        \includegraphics[width=\linewidth]{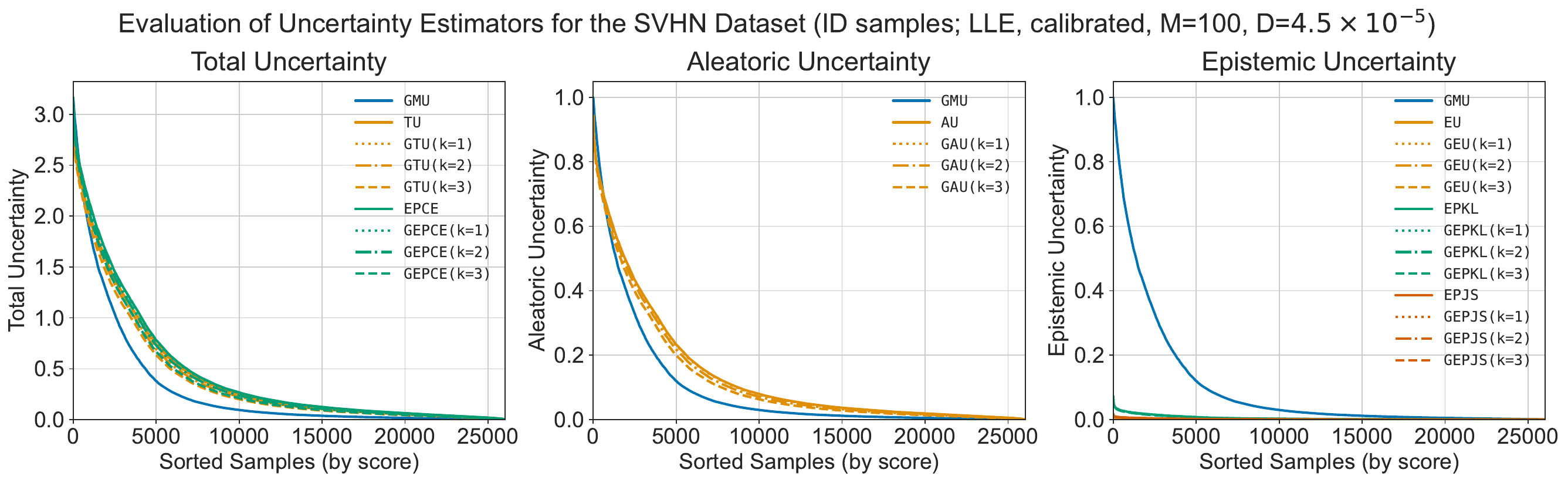}
    \end{subfigure}
    \begin{subfigure}{\linewidth}
        \centering
        \includegraphics[width=\linewidth]{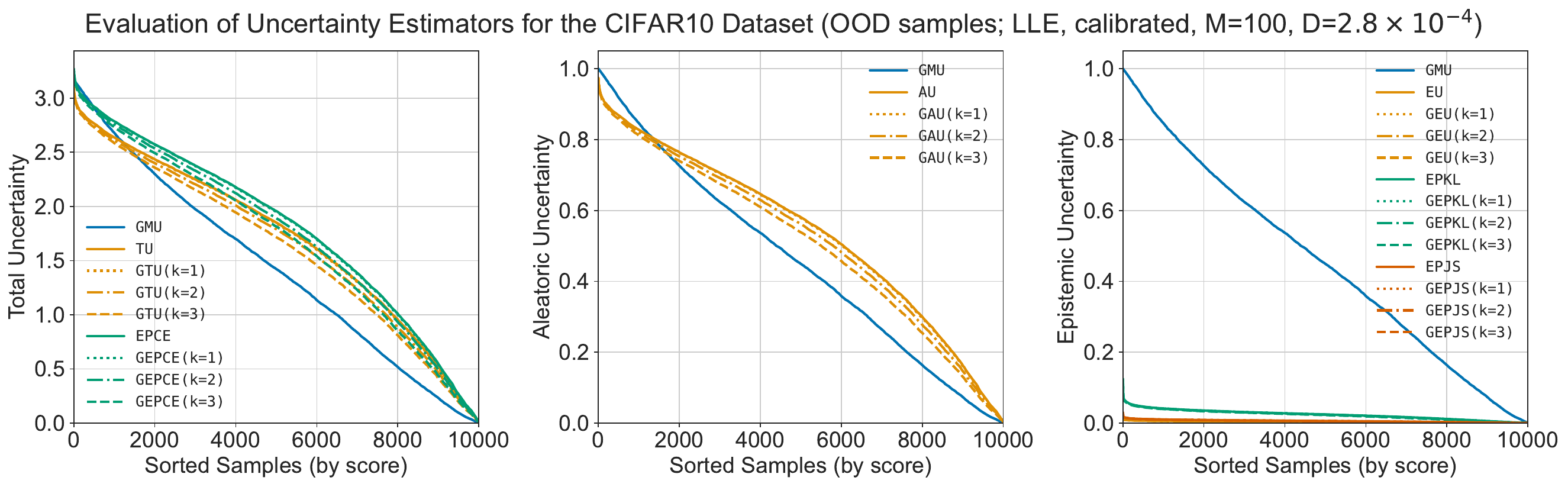}
    \end{subfigure}
    \begin{subfigure}{\linewidth}
        \centering
        \includegraphics[width=\linewidth]{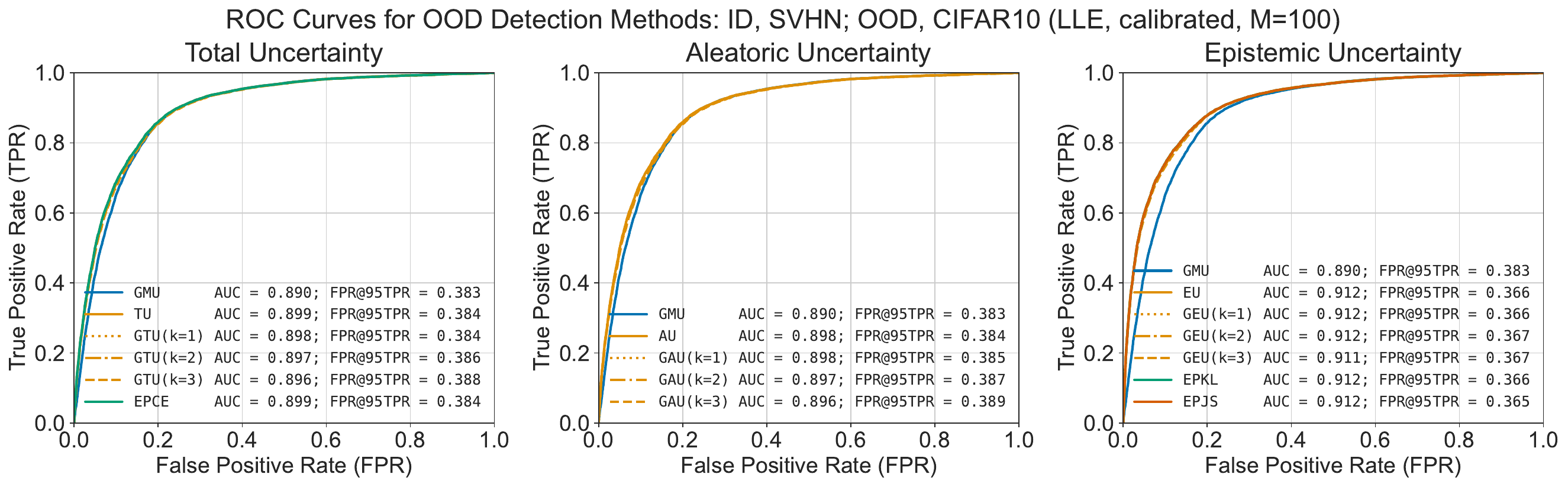}
    \end{subfigure}
    \caption{
    \textbf{Uncertainty decompositions} and \textbf{OOD} for the \textbf{SVHN} dataset using \textbf{LLE, calibrated} network and variance-gated distributions, compared against baseline measures.
    Diversity (D) was quantified as the expectation over samples $i$ and classes $c$, of the variance across models ($\mathbb{E}_{i,c}[\mathrm{Var}_M]$). 
    Uncertainty measures were normalized by the largest metric to allow direct comparison across methods
    }
\end{figure}

\clearpage

\subsection{CIFAR10}

\begin{figure}[!ht]
    \centering
    \begin{subfigure}{\linewidth}
        \centering
        \includegraphics[width=\linewidth]{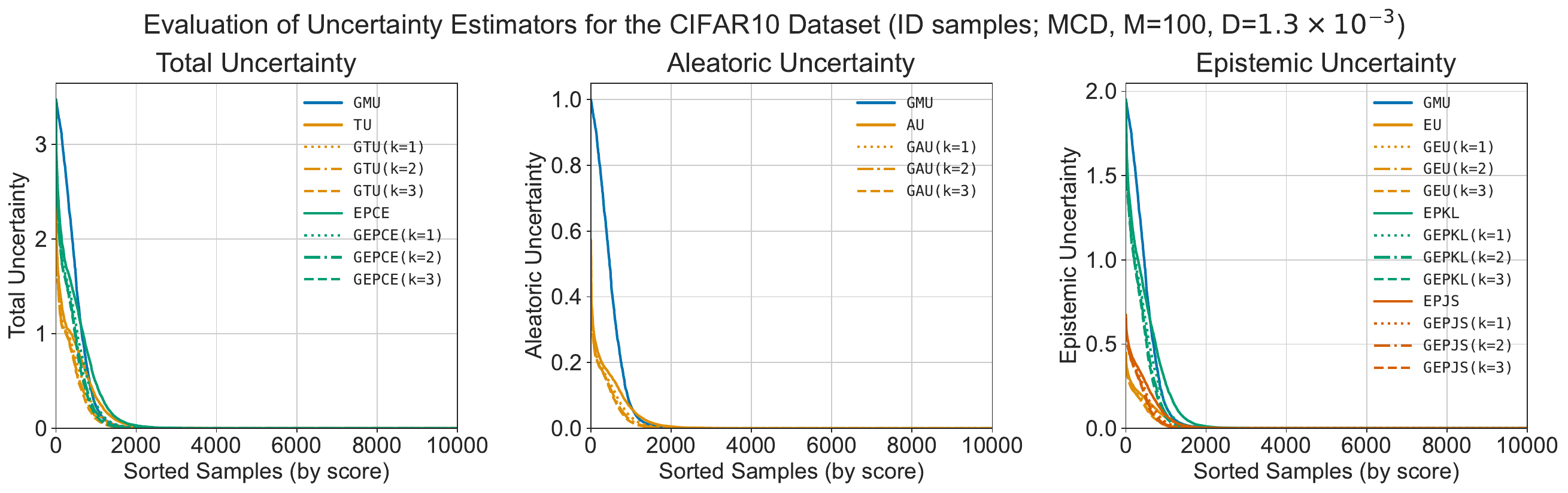}
    \end{subfigure}
    \begin{subfigure}{\linewidth}
        \centering
        \includegraphics[width=\linewidth]{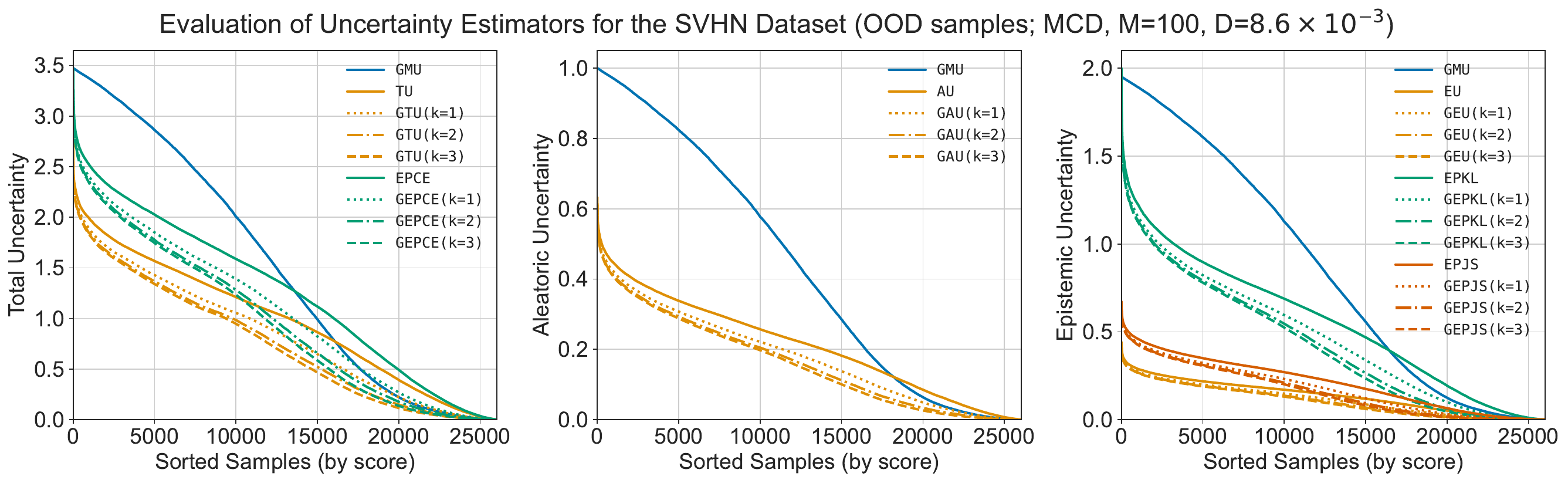}
    \end{subfigure}
    \begin{subfigure}{\linewidth}
        \centering
        \includegraphics[width=\linewidth]{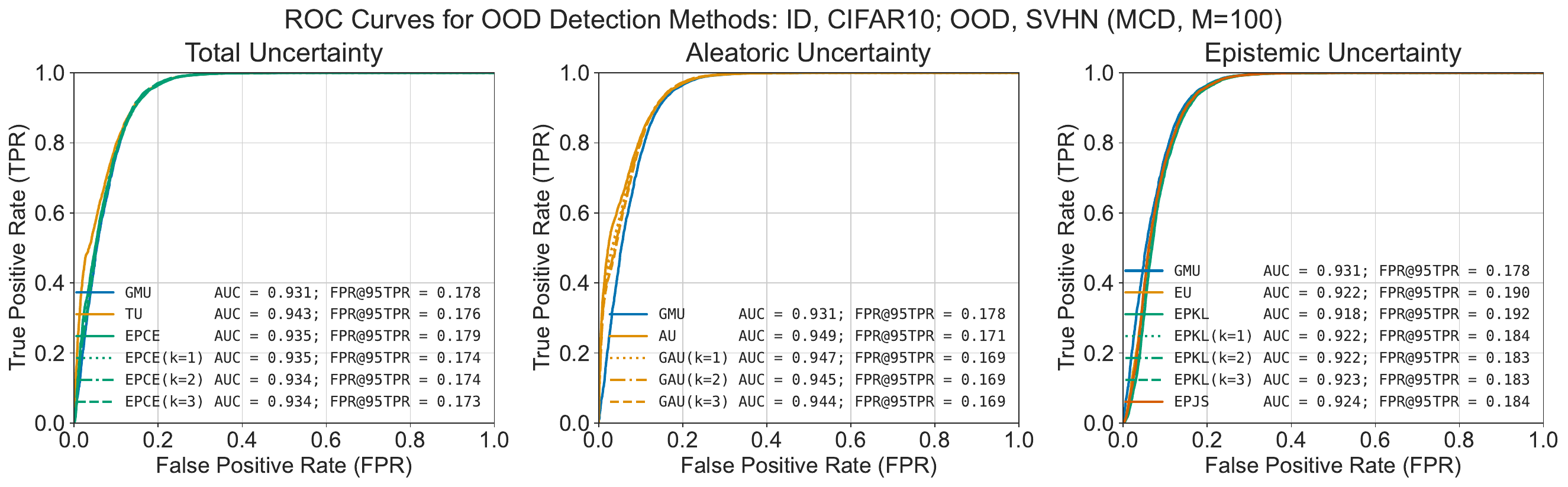}
    \end{subfigure}
    \caption{
    \textbf{Uncertainty decompositions} and \textbf{OOD} for the \textbf{CIFAR10} dataset using \textbf{MCD} network and variance-gated distributions, compared against baseline measures.
    Diversity (D) was quantified as the expectation over samples $i$ and classes $c$, of the variance across models ($\mathbb{E}_{i,c}[\mathrm{Var}_M]$). 
    Uncertainty measures were normalized by the largest metric to allow direct comparison across methods
    }
\end{figure}

\clearpage

\begin{figure}[!ht]
    \centering
    \begin{subfigure}{\linewidth}
        \centering
        \includegraphics[width=\linewidth]{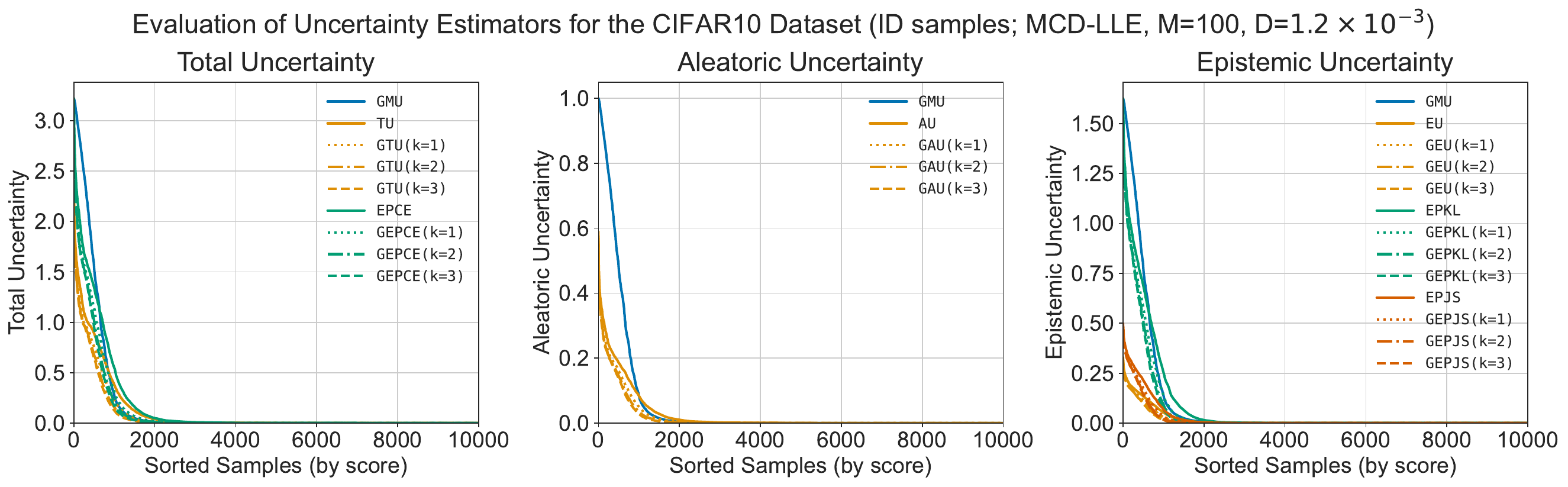}
    \end{subfigure}
    \begin{subfigure}{\linewidth}
        \centering
        \includegraphics[width=\linewidth]{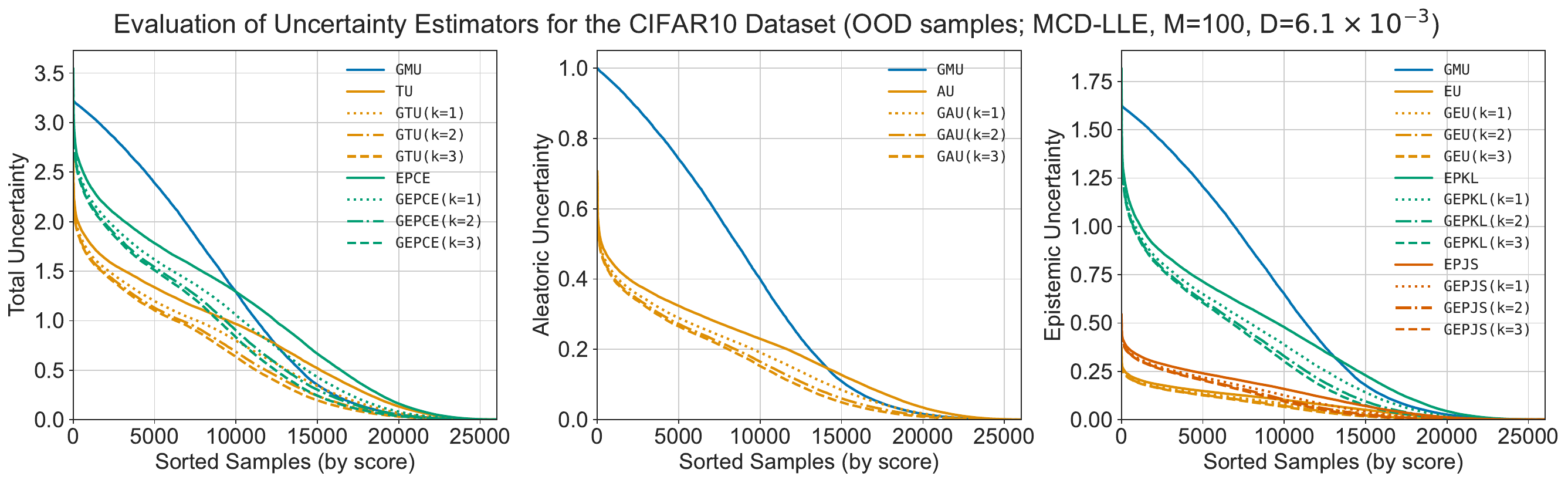}
    \end{subfigure}
    \begin{subfigure}{\linewidth}
        \centering
        \includegraphics[width=\linewidth]{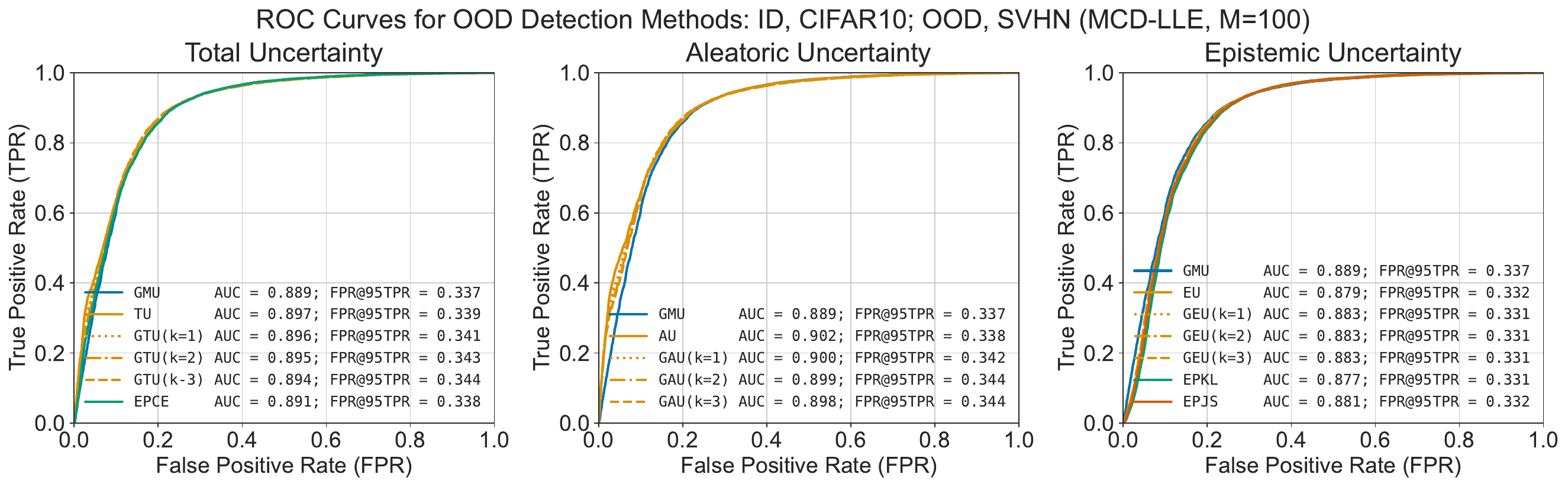}
    \end{subfigure}
    \caption{
    \textbf{Uncertainty decompositions} and \textbf{OOD} for the \textbf{CIFAR10} dataset using \textbf{MCD-LLE} network and variance-gated distributions, compared against baseline measures.
    Diversity (D) was quantified as the expectation over samples $i$ and classes $c$, of the variance across models ($\mathbb{E}_{i,c}[\mathrm{Var}_M]$). 
    Uncertainty measures were normalized by the largest metric to allow direct comparison across methods
    }
\end{figure}

\clearpage

\begin{figure}[!ht]
    \centering
    \begin{subfigure}{\linewidth}
        \centering
        \includegraphics[width=\linewidth]{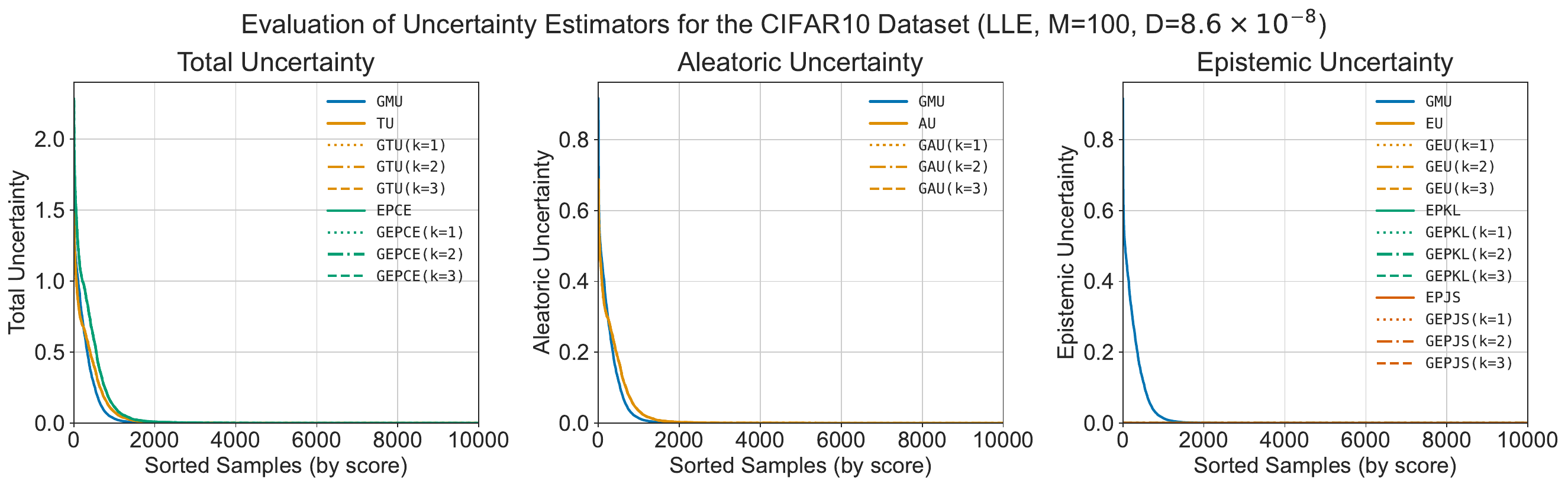}
    \end{subfigure}
    \begin{subfigure}{\linewidth}
        \centering
        \includegraphics[width=\linewidth]{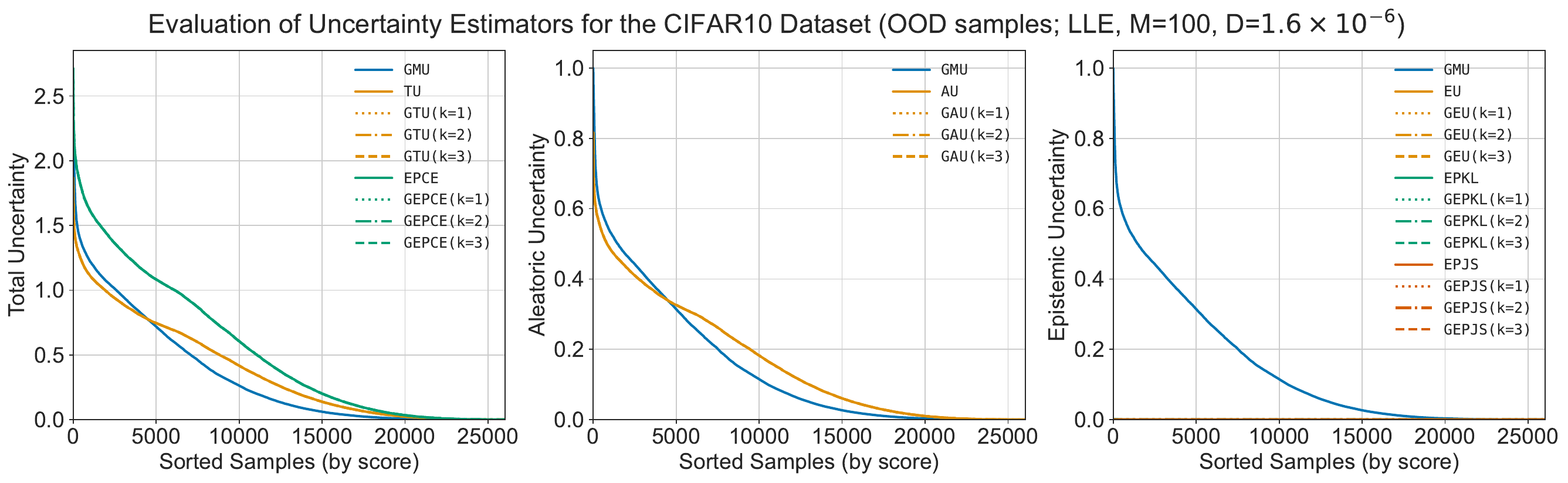}
    \end{subfigure}
    \begin{subfigure}{\linewidth}
        \centering
        \includegraphics[width=\linewidth]{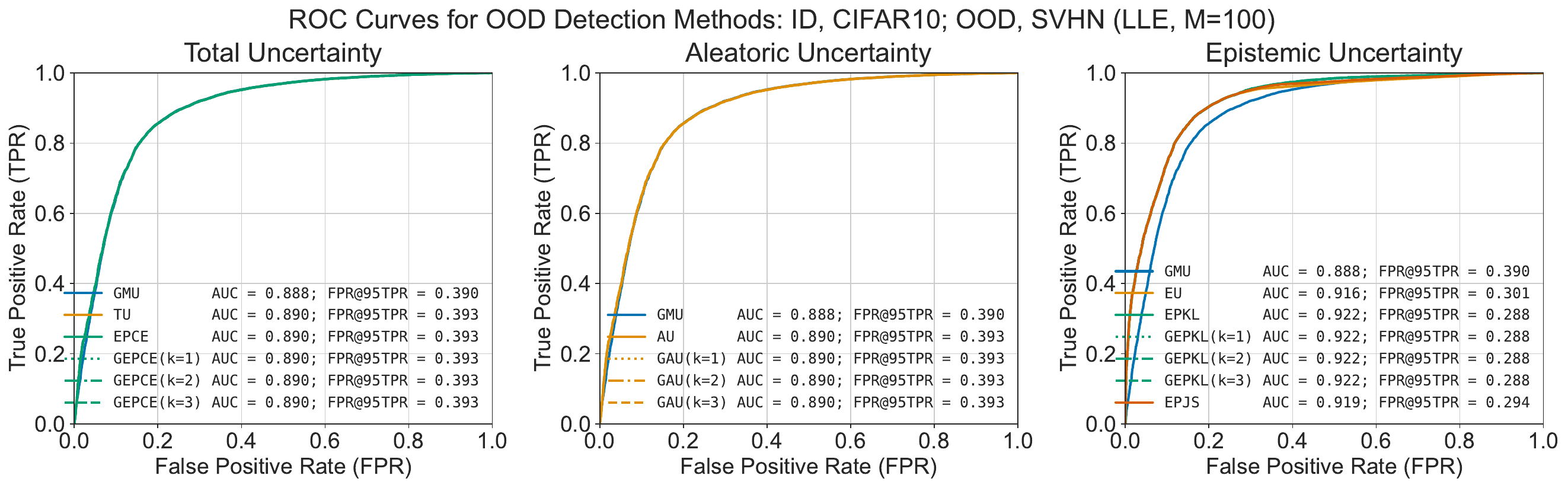}
    \end{subfigure}
    \caption{
    \textbf{Uncertainty decompositions} and \textbf{OOD} for the \textbf{CIFAR10} dataset using \textbf{LLE} network and variance-gated distributions, compared against baseline measures.
    Diversity (D) was quantified as the expectation over samples $i$ and classes $c$, of the variance across models ($\mathbb{E}_{i,c}[\mathrm{Var}_M]$). 
    Uncertainty measures were normalized by the largest metric to allow direct comparison across methods
    }
\end{figure}

\clearpage

\begin{figure}[!ht]
    \centering
    \begin{subfigure}{\linewidth}
        \centering
        \includegraphics[width=\linewidth]{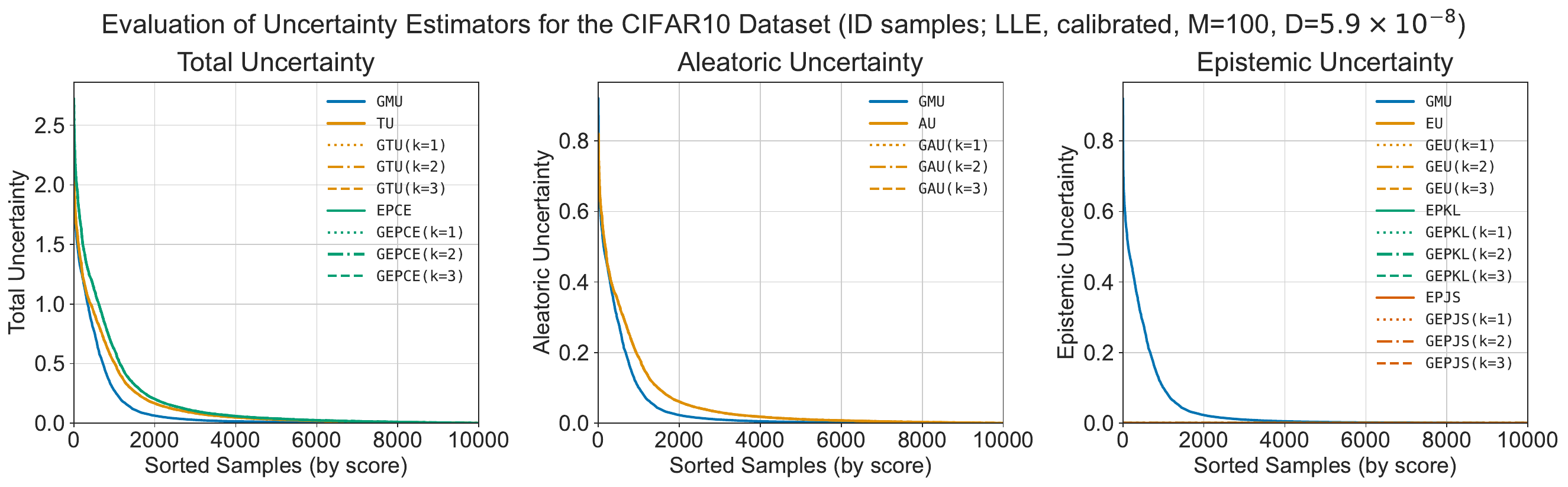}
    \end{subfigure}
    \begin{subfigure}{\linewidth}
        \centering
        \includegraphics[width=\linewidth]{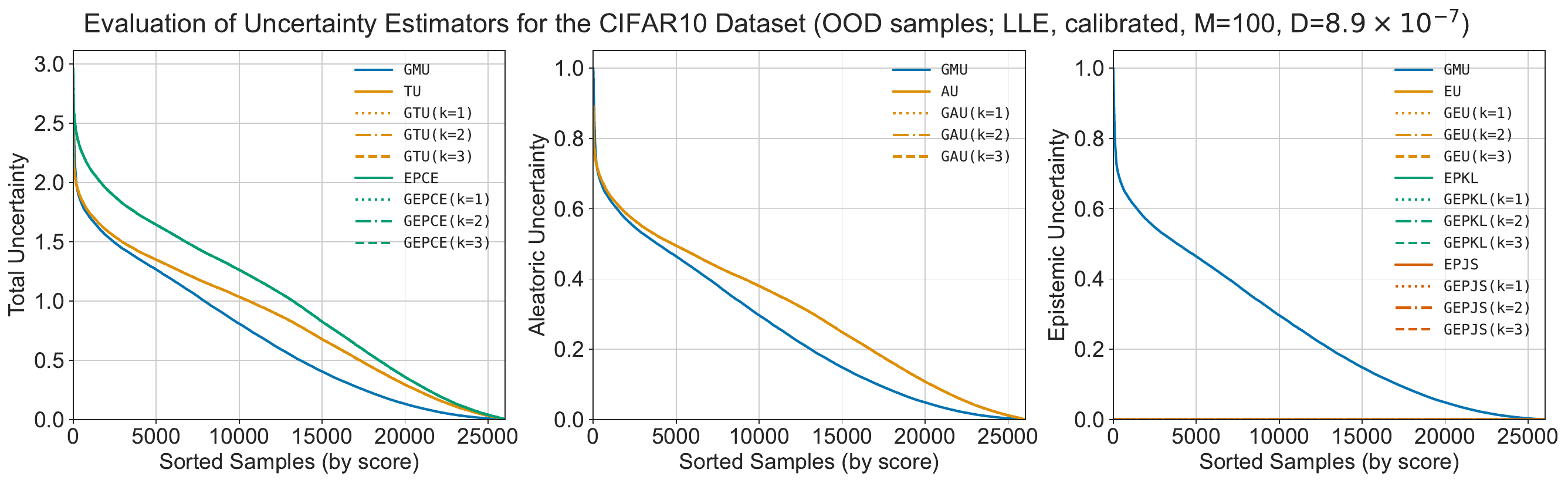}
    \end{subfigure}
    \begin{subfigure}{\linewidth}
        \centering
        \includegraphics[width=\linewidth]{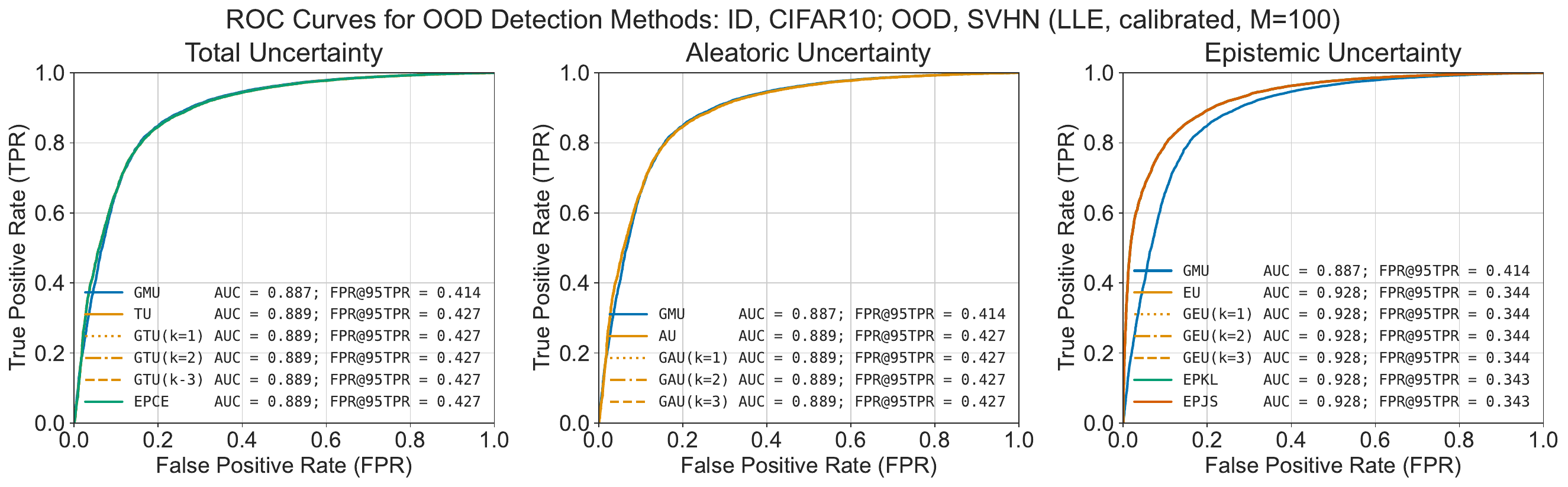}
    \end{subfigure}
    \caption{
    \textbf{Uncertainty decompositions} and \textbf{OOD} for the \textbf{CIFAR10} dataset using \textbf{LLE, calibrated} network and variance-gated distributions, compared against baseline measures.
    Diversity (D) was quantified as the expectation over samples $i$ and classes $c$, of the variance across models ($\mathbb{E}_{i,c}[\mathrm{Var}_M]$). 
    Uncertainty measures were normalized by the largest metric to allow direct comparison across methods
    }
\end{figure}

\clearpage

\subsection{CIFAR100}

\begin{figure}[!ht]
    \centering
    \begin{subfigure}{\linewidth}
        \centering
        \includegraphics[width=\linewidth]{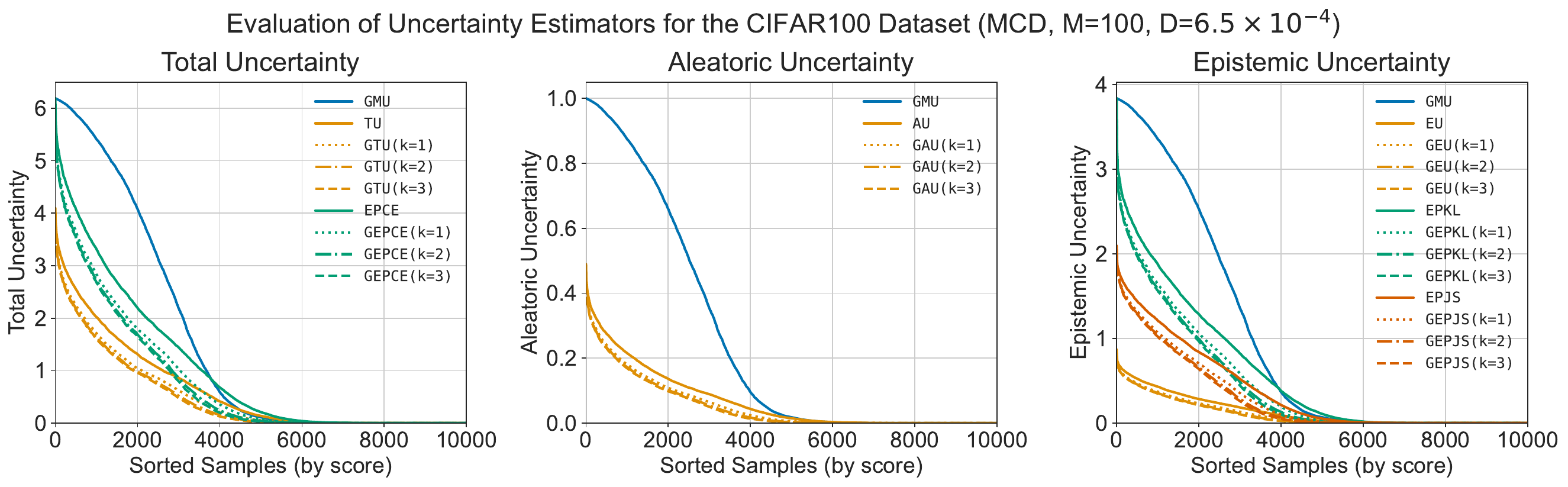}
    \end{subfigure}
    \begin{subfigure}{\linewidth}
        \centering
        \includegraphics[width=\linewidth]{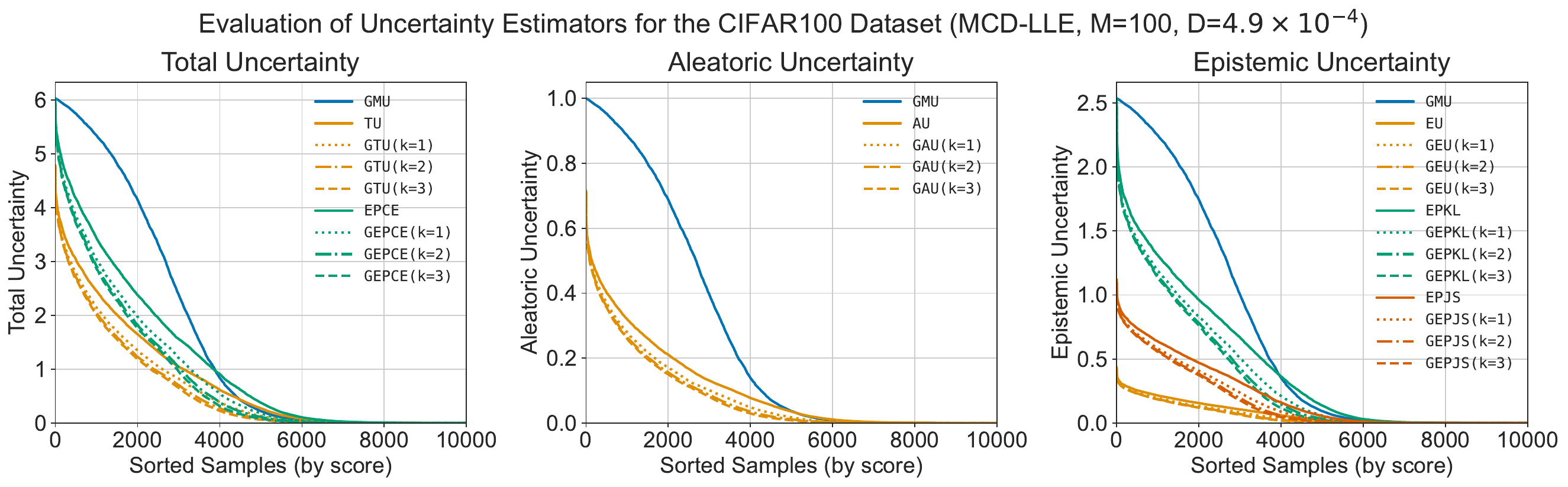}
    \end{subfigure}
    \begin{subfigure}{\linewidth}
        \centering
        \includegraphics[width=\linewidth]{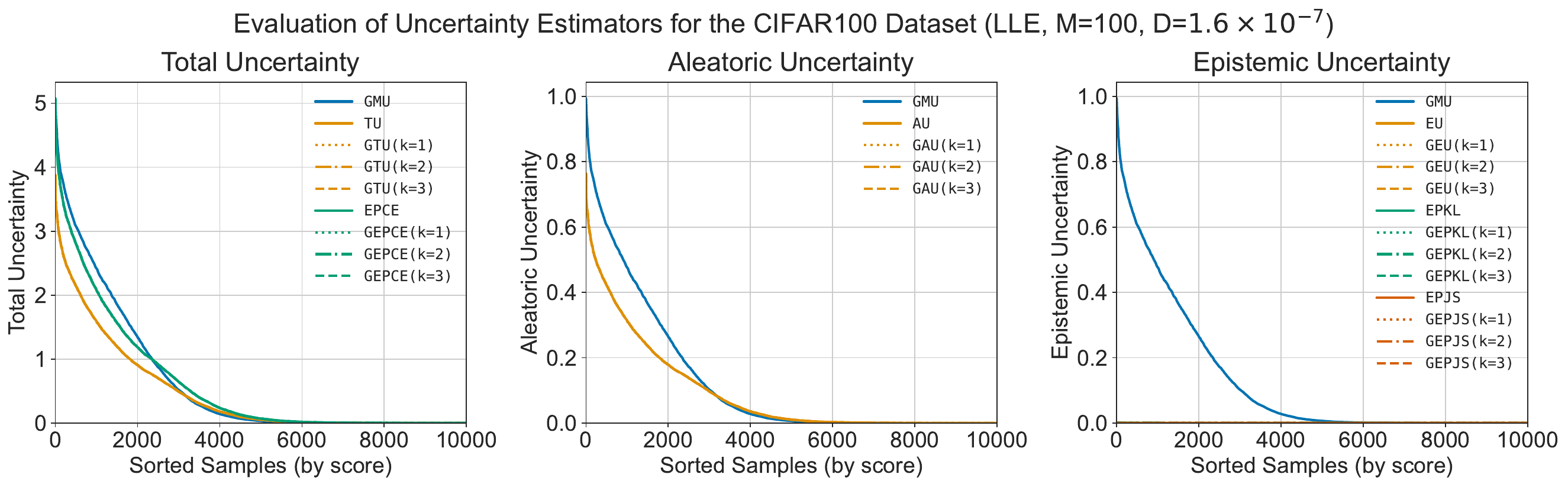}
    \end{subfigure}
    \begin{subfigure}{\linewidth}
        \centering
        \includegraphics[width=\linewidth]{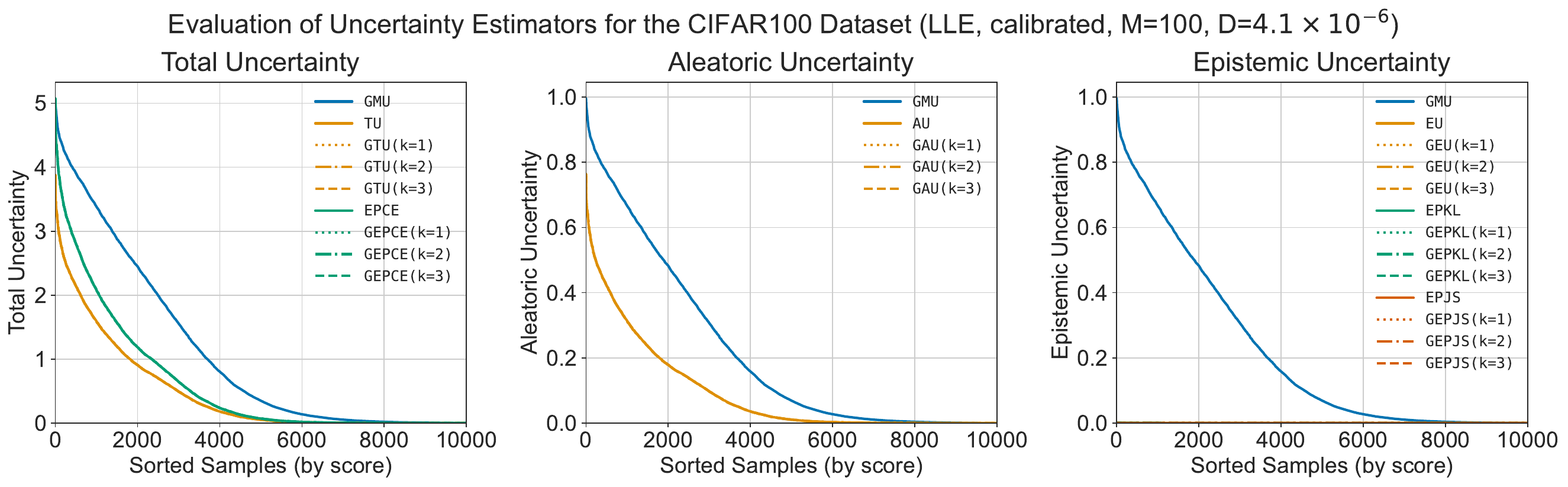}
    \end{subfigure}
    \caption{
    Uncertainty decompositions for the \textbf{CIFAR100} dataset using \textbf{MCD, MCD-LLE, LLE, and LLE (calibrated)} networks and variance-gated distributions, compared against baseline measures.
    Diversity (D) was quantified as the expectation over samples $i$ and classes $c$, of the variance across models ($\mathbb{E}_{i,c}[\mathrm{Var}_M]$). 
    Uncertainty measures were normalized by the largest metric to allow direct comparison across methods.
    }
\end{figure}

\clearpage

\end{document}